%% file: main.tex
\documentclass[10pt,twocolumn,letterpaper]{article}

\usepackage[pagenumbers]{wacv} 

\usepackage{booktabs}
\usepackage{multirow}
\usepackage{placeins} 
\usepackage{tabularx}
\usepackage{array}
\usepackage{amssymb}
\usepackage{tcolorbox}
\tcbuselibrary{breakable}
\usepackage{amsmath}
\usepackage{pifont}
\usepackage{titletoc}
\usepackage[accsupp]{axessibility}


\definecolor{wacvblue}{rgb}{0.21,0.49,0.74}
\usepackage[pagebackref,breaklinks,colorlinks,allcolors=wacvblue]{hyperref}

\def\wacvPaperID{425} 
\def\confName{WACV}
\def\confYear{2027}

\title{Humans in Space: Can MLLMs Construct Human-Centric Spatial Understanding via Active Exocentric Exploration?}

\author{
Tianchen Guo\textsuperscript{1,4}\quad
Tengyue Zhang\textsuperscript{2}\quad
Chen Liu\textsuperscript{3}\quad
Ling Chen\textsuperscript{3}\quad
Xin Yu\textsuperscript{2,4,$*$}\\[4pt]
\textsuperscript{1}The University of Queensland\quad
\textsuperscript{2}Australian Institute for Machine Learning, Adelaide University\\
\textsuperscript{3}University of Technology Sydney\quad
\textsuperscript{4}Follow Me AI Pty LTD
}

\begin{document}

\twocolumn[{%
\renewcommand\twocolumn[1][]{#1}%

\maketitle

\begin{center}
    \centering
    \captionsetup{hypcap=false}
    \includegraphics[width=0.92\textwidth]{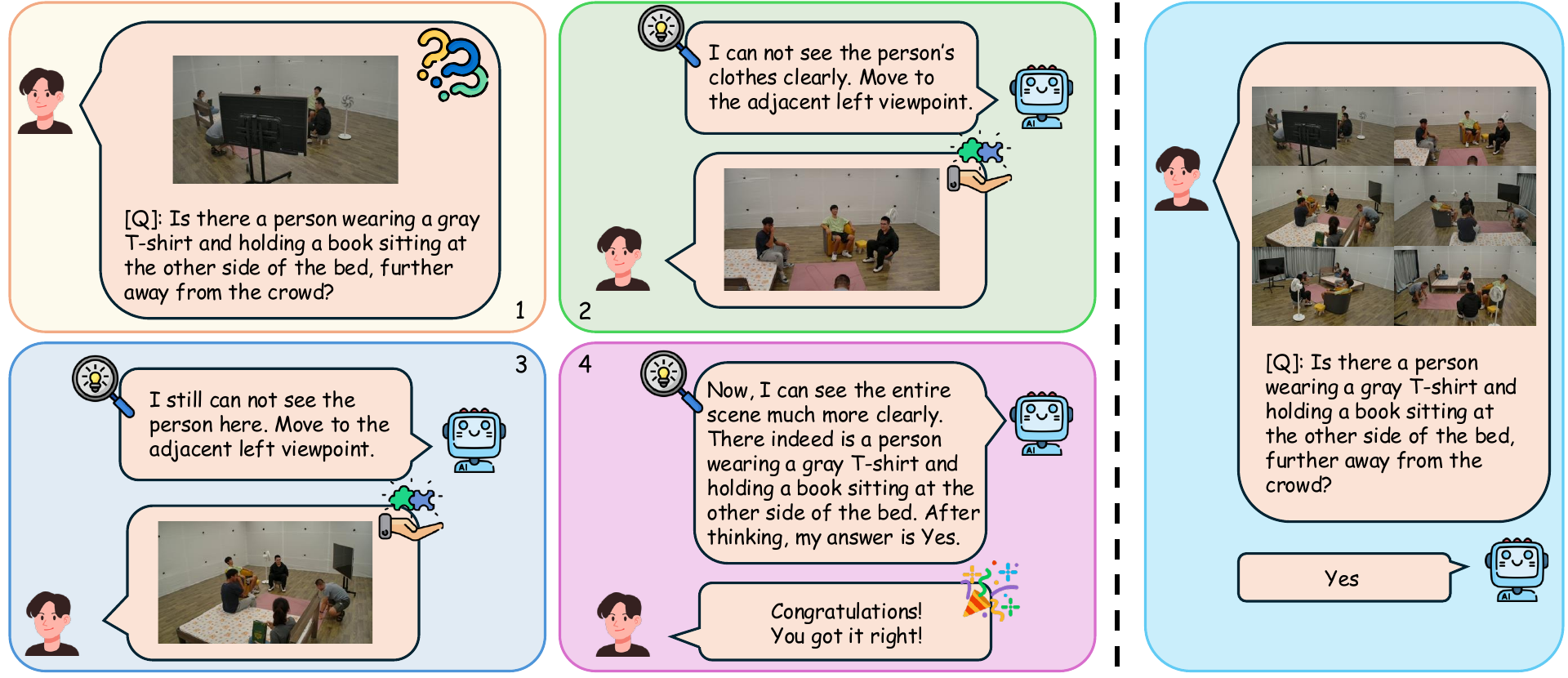}
    \vspace{-0.5em}
    \captionof{figure}{\textbf{Illustration of the SATE-Bench Evaluation.} SATE-Bench introduces a novel diagnostic framework to assess visual self-awareness in MLLMs. \textbf{Left (Active Setting):} The model is initialized with a restricted exocentric viewpoint where key information is either severely occluded or entirely outside the current field of view. Instead of guessing, it must recognize its visual uncertainty and iteratively query adjacent spatial perspectives (view-exploration) until it uncovers sufficient geometric evidence to answer the questions. \textbf{Right (Passive Setting):} The model is provided with multi-view images, testing pure fusion ability without the need for active exploration.}
    \label{fig:title}
\end{center}
}]%

{\let\thefootnote\relax\footnotetext{$^{*}$Corresponding author.}}
\input{sec/0_abstract}
\input{sec/1_full}

{
    \small
    \bibliographystyle{ieeenat_fullname}
    \bibliography{main}
}
\input{sec/X_suppl}

\end{document}

%% file: sec/0_abstract.tex
\begin{abstract}
While Multimodal Large Language Models (MLLMs) excel at zero-shot reasoning, their spatial capabilities are primarily evaluated through static, passive observation. In complex, human-centric scenes, occlusion often hides crucial geometric evidence, and passive benchmarks force models to guess rather than recognize their own visual uncertainty. To bridge this gap, we introduce \textbf{SATE-Bench}, a novel interactive benchmark comprising 316 real-world scenes and 2,525 multi-turn QA pairs to evaluate visual self-awareness and active exocentric view-exploration. It combines a Passive multi-view control with an Active exploration setting, where models start from a severely occluded view and must iteratively query adjacent viewpoints until they gather enough evidence to fully ``sate'' the question, scored by a step-penalized efficiency reward. Evaluating state-of-the-art MLLMs reveals that they terminate exploration prematurely while reporting high confidence, and struggle with fine-grained 3D geometric tasks; matched-view controls further attribute the Active--Passive gap primarily to insufficient evidence acquisition. Moreover, fine-tuning with Group Relative Policy Optimization (GRPO) shows that rewarding view-seeking increases active exploration but lowers overall accuracy, as models over-explore and fail to fuse expanding multi-view contexts across turns. These findings indicate that scaling passive multi-view perception does not intrinsically yield active spatial reasoning, highlighting the need for models that know where to look, how to integrate what they see, and when to stop. Code and data are available at \url{https://github.com/gtc-gh/SATE-Bench}.
\end{abstract}

%% file: sec/1_full.tex
\section{Introduction}
\label{sec:intro}
The visual perception of human behavior is fundamentally dynamic and multidimensional~\cite{locke1847essay, wang2025lvbench, yang2025vca, tang2025video, yeh2026seeing, guo2025beyond, qian2025dispider, zhang2025vlm2, hong2025motionbench, xun2026rtv,zhang2024affective,zhang2024effective,guo2024being,zhao2025pp,qiu2024language,qiu2024learning,liu2025dynamic,liu2025robust,liu2024benchmarking,song2023hairstyle,liu2023audio,liu2024bavs}. In real environments such as collaborative sports~\cite{van2022deepsportradar} and operating rooms~\cite{ozsoy2025mm}, severe mutual occlusion frequently obscures essential geometric cues, requiring an intelligent observer to actively seek alternative perspectives~\cite{wang2025actiview, khayatkhoei2025mllms, aloimonos1988active}. Despite the strong zero-shot reasoning capabilities of Multimodal Large Language Models (MLLMs)~\cite{achiam2023GPT, bai2025qwen3, wang2025internvl3, an2025llava, an2026llava}, current spatial evaluations remain primarily limited to passive observation. Standard Visual Question Answering (VQA) benchmarks~\cite{fu2024blink, wang2024muirbench,yeh2026seeing,yang2025thinking,fu2026mme,yin2025spatial,guo2025plnet} provide a predetermined set of images, thereby focusing on reasoning given that evidence while ignoring \textit{visual self-awareness}~\cite{wang2024mm}, \ie, the capacity to recognize observational blind spots. When faced with heavily occluded views, such passive frameworks force models to rely on linguistic priors and hallucinate missing details rather than seeking new information.


Existing interactive benchmarks~\cite{zhang2026theory,hong2026esi,wang2025actiview} predominantly focus on synthetic environments and ego-centric trajectory planning, leaving real-world, human-centric exocentric exploration largely unexamined. This setting is inherently challenging: a model must assess whether its current view is sufficient, decide which viewpoint to query next, and establish cross-view correspondence of people and objects across wide perspective shifts.
To bridge this gap, we introduce \textbf{SATE-Bench}. The acronym ``SATE'' reflects our core evaluation philosophy: to fully satisfy a desire. An initial occluded view leaves the model ``hungry'' for information, compelling it to actively explore the spatial environment to sate its need for definitive visual evidence. We construct this benchmark by sampling 316 dense, real-world scenes from 10 diverse multi-view human-centric datasets, yielding 2,525 high-quality QA pairs with an average of 6.04 camera views per scene.


To disentangle multi-view feature fusion from active ambiguity resolution, SATE-Bench pairs an Active exploration setting with a Passive control. In the Active setting, the MLLM is initialized with a single, severely occluded view and must autonomously judge whether it has sufficient evidence to answer; if not, it iteratively queries adjacent exocentric viewpoints (left or right) across multiple turns until the necessary spatial information is uncovered. Conversely, the Passive setting isolates pure multi-view spatial reasoning by providing a comprehensive set of static views upfront, guaranteeing that the question is directly answerable without exploration. Comparing performance across both settings tests whether a model knows when to seek new views and when it has gathered enough evidence to stop.

\begin{table}[t]
\centering
\caption{Comprehensive comparison of spatial reasoning and visual perception benchmarks. Abbreviations: \textbf{Act.} (Active), \textbf{Pas.} (Passive), \textbf{Real} (Real-world scenes), \textbf{MT} (Multi-Turn, supporting interaction trajectories of $>3$ consecutive turns), \textbf{DEMV} (Dense Exocentric Multi-View), \textbf{SA} (Self-Awareness), \textbf{SO} (Severe Occlusion), \textbf{HC} (Human-Centric).}
\label{tab:benchmark_comparison}
\small
\setlength{\tabcolsep}{3pt}
\resizebox{\columnwidth}{!}{
\begin{tabular}{l|cccccccc}
\toprule
\textbf{Benchmark} & \textbf{Act.} & \textbf{Pas.} & \textbf{Real} & \textbf{MT} & \textbf{DEMV} & \textbf{SA} & \textbf{SO} & \textbf{HC} \\
\midrule
\textbf{BLINK~\cite{fu2024blink}}          & \ding{55} & \ding{51} & \ding{51} & \ding{55} & \ding{55} & \ding{55} & \ding{55} & \ding{55} \\
\textbf{MuirBench~\cite{wang2024muirbench}}      & \ding{55} & \ding{51} & \ding{51} & \ding{55} & \ding{55} & \ding{55} & \ding{55} & \ding{55} \\
\textbf{MMSI-Bench~\cite{yang2025mmsi}}      & \ding{55} & \ding{51} & \ding{51} & \ding{55} & \ding{55} & \ding{55} & \ding{55} & \ding{55} \\
\textbf{AllAnglesBench~\cite{yeh2026seeing}} & \ding{55} & \ding{51} & \ding{51} & \ding{55} & \ding{55} & \ding{55} & \ding{55} & \ding{55} \\
\textbf{MindCube~\cite{yin2025spatial}}       & \ding{55} & \ding{51} & \ding{51} & \ding{55} & \ding{55} & \ding{55} & \ding{51} & \ding{55} \\
\textbf{CVBench~\cite{guo2026beyond}}        & \ding{55} & \ding{51} & \ding{51} & \ding{55} & \ding{55} & \ding{55} & \ding{55} & \ding{51} \\
\textbf{ActiView~\cite{wang2025actiview}}       & \ding{51} & \ding{55} & \ding{51} & \ding{55} & \ding{55} & \ding{55} & \ding{55} & \ding{55} \\
\textbf{Theory of Space~\cite{zhang2026theory}}& \ding{51} & \ding{51} & \ding{55} & \ding{51} & \ding{55} & \ding{51} & \ding{55} & \ding{55} \\
\textbf{ProactiveBench~\cite{de2026proactivebench}} & \ding{51} & \ding{51} & \ding{51} & \ding{55} & \ding{55} & \ding{55} & \ding{55} & \ding{55} \\
\textbf{ESI-Bench~\cite{hong2026esi}}      & \ding{51} & \ding{51} & \ding{55} & \ding{51} & \ding{55} & \ding{51} & \ding{51} & \ding{55} \\
\midrule
\textbf{SATE-Bench} (Ours) & \ding{51} & \ding{51} & \ding{51} & \ding{51} & \ding{51} & \ding{51} & \ding{51} & \ding{51} \\
\bottomrule
\end{tabular}
}
\end{table}

\begin{figure*}[tb]
  \centering
  \includegraphics[width=0.9\textwidth]{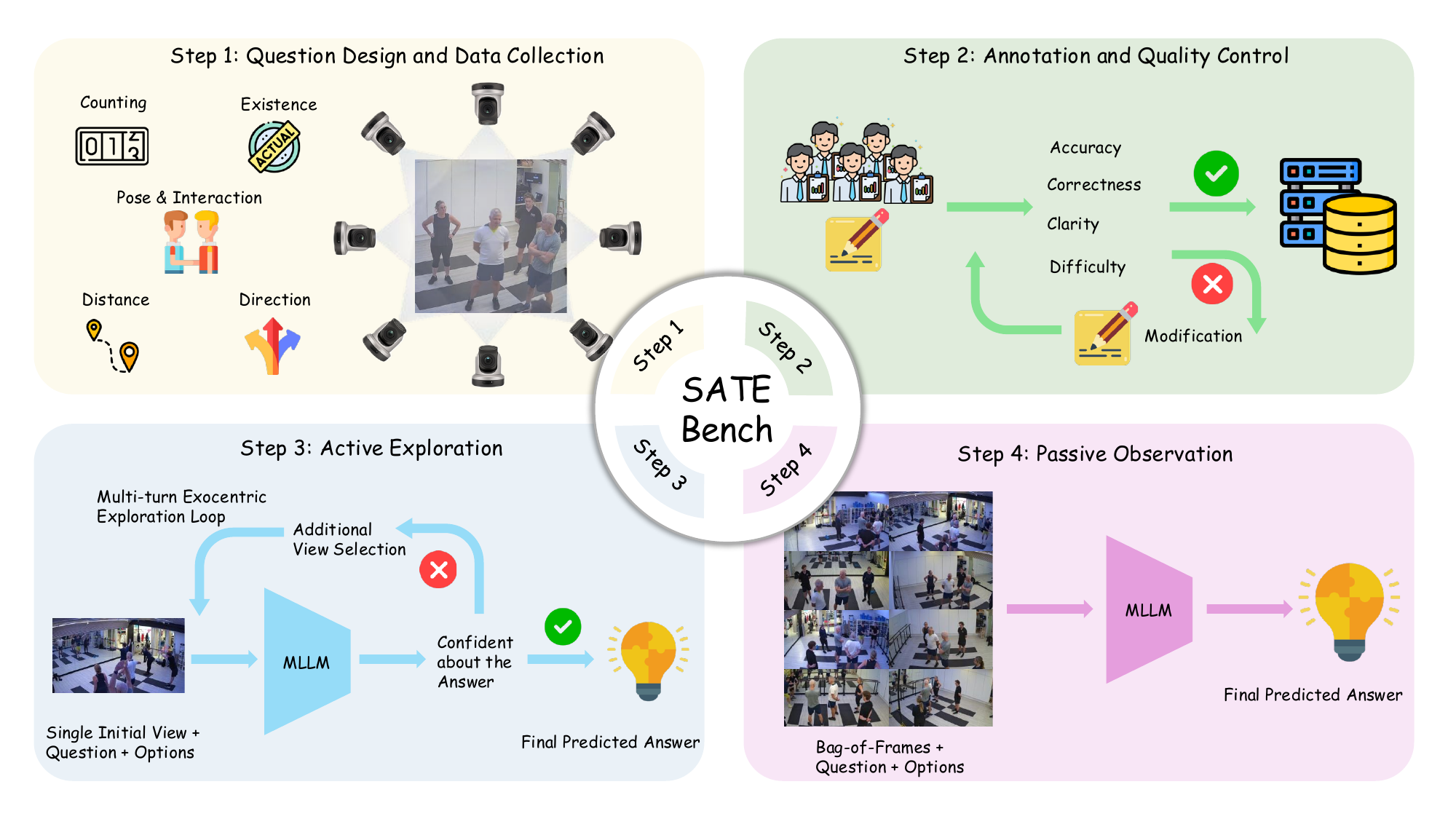}
  \caption{\textbf{Overview of the SATE-Bench Curation and Evaluation Pipeline.} Our framework is constructed and deployed through four distinct stages. \textbf{Steps 1 \& 2 (Data Collection and Quality Control):} We source dense, 360-degree multi-view captures across five cognitive spatial categories. Candidate question-answer pairs undergo a rigorous human annotation and modification loop to guarantee accuracy, clarity, and appropriate difficulty. \textbf{Step 3 (Active Exploration):} The multi-turn interactive evaluation setting. The MLLM is initialized with a single view and must dynamically query additional exocentric perspectives until it confidently resolves its visual uncertainty. \textbf{Step 4 (Passive Observation):} The single-turn control setting, where the model is provided a static, comprehensive bag-of-frames to yield a direct prediction without the need for spatial view-exploration.}
  \label{fig:teaser}
\end{figure*}

SATE-Bench also incorporates an exponentially decaying reward to penalize redundant queries and quantify exploration efficiency. Benchmarking 12 state-of-the-art MLLMs reveals that all models drop by 5--26\% in the Active setting, primarily due to overconfident, premature stopping: for most models, over 70\% of Active errors occur at the initial view without a single move. 
Moreover, applying Group Relative Policy Optimization (GRPO) to Qwen2.5-VL-3B shows that rewarding view-seeking boosts its active exploration rate from 0.2\% to 72.8\%, yet reduces accuracy from 30.4\% to 17.9\%, as the model moves more without learning when to stop or how to fuse the accumulated views. 


In summary, our main contributions are as follows:

\begin{itemize}
    \item \textbf{Novel Evaluation Protocol:} We identify the limitations of passive evaluation and formulate a diagnostic Active/Passive evaluation framework paired with an exponentially decaying efficiency reward to assess visual self-awareness and penalize redundant spatial queries.
    \item \textbf{SATE-Bench:} We curate an interactive benchmark of 316 real-world scenes and 2,525 multi-turn QA pairs for evaluating active exocentric view-exploration.
    \item \textbf{Comprehensive Benchmarking:} We evaluate state-of-the-art MLLMs, showing through confidence probing and matched-view controls that premature stopping and insufficient evidence acquisition drive the Active--Passive gap.
    \item \textbf{Active Exploration via GRPO:} We show that GRPO boosts active view-seeking while uncovering a trade-off between active exploration and reasoning accuracy.
\end{itemize}

\section{Related Work}


\subsection{Multi-Image MLLM Benchmarks}
Recent multi-image MLLM benchmarks universally evaluate reasoning via static, passive observation~\cite{yang2025thinking, gholami2026spatial, wang2024muirbench, fu2024blink, yang2025mmsi, li2024mvbench, yue2024mmmu, wang2022mvp, plizzari2025omnia, xun2026rtv, zhou2025urbench, liu2024mmbench}. Although recent multi-view evaluations like AllAnglesBench~\cite{yeh2026seeing} and CVBench~\cite{guo2026beyond} explore cross-view correspondence and spatial perception, they rely on pre-configured, static viewpoints. These benchmarks provide sparse perspective coverage (averaging just 3 views per scene in AllAnglesBench and 4 in CVBench) and fundamentally lack the interactive mechanics required to evaluate a model's active exploration abilities. When severe occlusion obscures important geometry, such passive frameworks force models to guess or hallucinate missing details rather than actively resolving visual ambiguity~\cite{li2023evaluating, bao2025faithbench}. While interactive setups like ActiView~\cite{wang2025actiview} permit 2D image cropping, they lack real-world 3D spatial complexity. 
Unlike prior benchmarks, as discussed in Table~\ref{tab:benchmark_comparison}, that rely on passive multi-view aggregation, SATE-Bench requires active spatial reasoning by situating MLLMs within immersive 360-degree environments to plan exocentric viewpoints and resolve visual occlusions.

\subsection{Active Visual Perception}

Active visual perception compels intelligent observers to dynamically adjust viewpoints to resolve uncertainty~\cite{aloimonos1988active, findlay2003active, wang2024mm}. While recent benchmarks like Theory of Space~\cite{zhang2026theory} and ESI-Bench~\cite{hong2026esi} advance active spatial reasoning, they focus predominantly on egocentric navigation within synthetic simulations to evaluate cognitive map construction and physical object manipulation. Similarly, ProactiveBench~\cite{de2026proactivebench} assesses proactive querying behaviors but lacks complex multi-view spatial environments. In contrast, \textbf{SATE-Bench} targets active exocentric view-exploration in dense, real-world human-centric scenes. Rather than assessing embodied navigation or general proactivity, it isolates \textit{visual self-awareness} under severe mutual occlusion, evaluating whether models can recognize observational blind spots and integrate cross-view geometry through a step-penalized, Active/Passive diagnostic framework.

\subsection{Multi-View Human-Centric Datasets}
Comprehending complex human behavior requires robust multi-view observations to overcome severe self- and mutual-occlusion. The community has developed diverse datasets capturing dense activities (M3GYM~\cite{xu2025m3gym}, Harmony4D~\cite{khirodkar2024harmony4d}, EgoHumans~\cite{khirodkar2023ego}), fine-grained interactions (HOI-M3~\cite{zhang2024hoi}, Ego-Exo4D~\cite{grauman2024ego}, core4D~\cite{liu2025core4d}), and occluded workflows (4D-OR~\cite{ozsoy20224d}, MM-OR~\cite{ozsoy2025mm}). While highly effective for advancing passive spatial tasks like 2D/3D pose estimation, human tracking, and human-object interaction detection, these datasets remain underutilized for interactive multi-turn conversational foundation model evaluation. SATE-Bench reuses these static multi-view captures into dynamic exploratory arenas, challenging MLLMs to autonomously step through discrete spatial camera views to uncover obscured visual truths.

\section{SATE-Bench}

\subsection{Benchmark Overview}
Humans inherently resolve spatial occlusions by actively exploring alternative viewpoints and internally fusing these multi-perspective observations. Yet, existing benchmarks~\cite{yeh2026seeing, guo2026beyond, hong2026esi, de2026proactivebench, zhang2026theory, antol2015vqa, goyal2017making, kazemi2024remi} primarily rely on single ego-centric views or passive, static image sets, leaving exocentric view-exploration and visual self-awareness largely unexamined. To bridge this gap, we introduce \textbf{SATE-Bench}. Rather than providing a predetermined ``bag of frames'', we situate the model within an interactive 360-degree spatial environment. Initialized with a single, heavily occluded view, the MLLM must explicitly recognize its visual uncertainty and dynamically query new exocentric perspectives to gather missing evidence. Through our decoupled framework of \textit{Active Exploration} and \textit{Passive Observation}, SATE-Bench evaluates whether an MLLM possesses the visual self-awareness to know what it cannot see, when to iteratively explore, and when to confidently answer. The overall framework is shown in Figure~\ref{fig:title} and Figure~\ref{fig:teaser}.

\subsection{Benchmark Curation Process}

\subsubsection{Task Definitions}
We define ten human-centric tasks across five cognitive dimensions to systematically evaluate spatial reasoning and view-exploration. \textbf{Entity Counting} (Human/Object) tests global spatial memory and cross-view re-identification, requiring models to resolve occlusions without double-counting. \textbf{Entity Existence} (Human/Object) probes visual self-awareness and blind-spot recognition by challenging models to query new viewpoints to confirm target presence. \textbf{Pose and Physical Interaction} (Action \& Pose / Human-Object Interaction) assesses granular geometric alignment and the resolution of severe self-occlusions. \textbf{Global Relative Proximity} (Distance to Human/Object) evaluates multi-view depth estimation and cross-perspective distance reasoning. Finally, \textbf{View-Conditioned Spatial Geometry} (Depth Ordering / Direction) tests perspective-taking, geometric projection, and dynamic coordinate frame alignment from specific, non-ego vantage points.

\begin{table}[tb]
\centering
\caption{Dataset Statistics and View Distributions by Task. ``Avail. Views'' and ``Passive Views'' denote the average number of available views and selected passive views, respectively.}
\label{tab:dataset_statistics}
\resizebox{\linewidth}{!}{
\begin{tabular}{lccc}
\toprule
\textbf{Question Name} & \textbf{\# QA Pairs} & \textbf{Avail. Views} & \textbf{Passive Views} \\
\midrule
Fine-grained Action and Pose & 301 & 5.23 & 2.22 \\
Human Counting               & 308 & 6.41 & 3.27 \\
Human Existence              & 203 & 7.57 & 3.33 \\
Human-Object Interaction     & 165 & 6.79 & 2.75 \\
Object Counting              & 232 & 5.17 & 3.46 \\
Object Existence             & 218 & 4.06 & 3.71 \\
Relative Distance to Human   & 508 & 5.09 & 2.21 \\
Relative Distance to Object  & 334 & 4.75 & 2.14 \\
View-Specific Depth Ordering & 124 & 7.21 & 3.46 \\
View-Specific Direction      & 132 & 7.92 & 4.08 \\
\midrule
\multicolumn{4}{p{11.5cm}}{\textbf{Data Sources:} M3GYM, Harmony4D, HOI-M3, MvMHAT, core4D, 4D-OR, Replay, EgoHumans, Ego-Exo4D, MM-OR. \hfill \textbf{Total QA Pairs: 2525}} \\
\bottomrule
\end{tabular}
}
\end{table}

\begin{table}[tb]
\centering
\caption{Dataset Text Length Statistics and Answer Distribution.}
\label{tab:text_statistics}
\resizebox{\linewidth}{!}{%
\begin{tabular}[t]{lr}
\toprule
\multicolumn{2}{c}{\textbf{Text Length Statistics}} \\
\midrule
\textbf{Metric} & \textbf{Value} \\
\midrule
Average Question Length & 102.89 \\
Max Question Length & 253.00 \\
Average Option Length & 49.33 \\
Max Option Length & 148.00 \\
Average `Purely Text' Option Length & 60.67 \\
Average Correct Answer Length & 33.66 \\
Max Correct Answer Length & 148.00 \\
\bottomrule
\end{tabular}%
\hspace{0.5em}%
\begin{tabular}[t]{lr}
\toprule
\multicolumn{2}{c}{\textbf{Answer Distribution}} \\
\midrule
\textbf{Option} & \textbf{Count} \\
\midrule
Option A & 505 \\
Option B & 505 \\
Option C & 505 \\
Option D & 505 \\
Option E & 505 \\
\bottomrule
\end{tabular}%
}
\end{table}

\subsubsection{Data Collection}
We source raw multi-view video sequences from diverse, real-world human-centric datasets, including M3GYM~\cite{xu2025m3gym}, Harmony4D~\cite{khirodkar2024harmony4d}, HOI-M3~\cite{zhang2024hoi}, Ego-Exo4D~\cite{grauman2024ego}, EgoHumans~\cite{khirodkar2023ego}, MvMHAT~\cite{gan2021mvmhat}, Replay~\cite{shapovalov2023replay}, 4D-OR~\cite{ozsoy20224d}, MM-OR~\cite{ozsoy2025mm}, and core4D~\cite{liu2025core4d}. These datasets are specifically selected for their dense physical overlap, multi-person interactions, and severe mutual/self-occlusion, ensuring that cross-view exploration is genuinely required to comprehend the scenes.

\subsubsection{Question-Answer Annotations}
\noindent\textbf{Annotation Process.} Four computer science annotators underwent an 8-hour training on 100 samples. They examined synchronized frames to create challenging multiple-choice QA pairs. For the Active setting, annotators selected initial occluded views that explicitly require view-exploration to gather missing information. For the Passive setting, they selected a subset of views guaranteeing full spatial coverage and definitive answerability. To prevent bias toward specific viewpoints or human preferences, these active and passive views were randomly sampled by the annotators from all valid camera configurations satisfying our task definitions.

\subsubsection{Quality Control}
To ensure logical soundness and visual necessity, two independent reviewers cross-validated the data, requiring $\ge 85\%$ agreement on the correct answer and initial active view, with a third expert resolving disputes. Reviewers removed ambiguous items and those answerable via pure text. Ultimately, 9.2\% of candidate items were discarded, and 38.7\% were revised for clarity, supplemented by automated syntax enhancement.




\begin{table*}[tb]
\centering
\caption{Comprehensive Evaluation Results on SATE-Bench across both Active and Passive settings. Task abbreviations: \textbf{HC} (Human Counting), \textbf{OC} (Object Counting), \textbf{FAP} (Fine-grained Action \& Pose), \textbf{HOI} (Human-Object Interaction), \textbf{HE} (Human Existence), \textbf{OE} (Object Existence), \textbf{RD-H} (Relative Distance to Human), \textbf{RD-O} (Relative Distance to Object), \textbf{VDO} (View-Specific Depth Ordering), \textbf{VSD} (View-Specific Direction). In the Passive setting, metrics requiring multi-turn exploration (Rewards, Steps, Active Rate) are marked with ``$-$''. For enhanced readability, the reported Average Reward is scaled by $100$. $^*$Evaluated based on Gemini-3.1-flash-lite. Best results among models are bolded. More open-source model results can be found in the Appendix~\ref{app:more_open_source}.}
\label{tab:main_results}
\resizebox{\textwidth}{!}{%
\begin{tabular}{l|cccccccccc|ccccc}
\toprule
\textbf{Model} & \textbf{HC} & \textbf{OC} & \textbf{FAP} & \textbf{HOI} & \textbf{HE} & \textbf{OE} & \textbf{RD-H} & \textbf{RD-O} & \textbf{VDO} & \textbf{VSD} & \textbf{Avg.} & \textbf{Avg.} & \textbf{Avg.} & \textbf{Active} & \textbf{Inf. Time} \\
& & & & & & & & & & & \textbf{Acc (\%)} & \textbf{Steps} & \textbf{Reward} & \textbf{Rate (\%)} & \textbf{(hours)} \\
\midrule
\multicolumn{16}{l}{\textit{Baselines}} \\
\midrule
Random Guess          & 18.2 & 22.8 & 19.5 & 15.6 & 50.5 & 54.0 & 22.1 & 16.7 & 21.3 & 21.9 & 24.8 & 1.0 & 22.5 & 0.0 & 0.0 \\
Blind (Text-only)     & 0.4 & 6.3 & 20.5 & 3.3 & 46.8 & 53.2 & 21.5 & 21.7 & 20.0 & 31.2 & 20.9 & 2.1 & 19.1 & 21.5 & 1.9 \\
Human  & 88.2 & 76.8 & 79.6 & 83.1 & 81.4 & 87.6 & 78.9 & 87.1 & 80.7 & 84.3 & 82.7 & 4.7 & 55.6 & 88.1 & $>$10.0 \\
Oracle-View$^*$          & 42.2 & 26.3 & 67.9 & 17.1 & 60.4 & 57.4 & 73.5 & 59.7 & 25.8 & 33.3 & 54.7 & -   & -    & -  & $<$1.0  \\
Oracle-Trajectory$^*$    & 37.3 & 28.9 & 61.3 & 31.7 & 58.5 & 57.4 & 74.0 & 64.3 & 29.0 & 30.6 & 55.2 & -   & -    & -  & $<$1.0  \\
Always Explore Left$^*$  & 35.6 & 25.2 & 55.9 & 30.0 & 68.5 & 50.0 & 48.7 & 45.6 & 18.7 & 37.5 & 43.6 & 2.0 & 36.5 & 44.8 & 1.4 \\
Always Explore Right$^*$ & 32.9 & 27.6 & 55.9 & 32.2 & 67.6 & 44.4 & 47.1 & 39.4 & 24.0 & 39.6 & 42.4 & 1.9 & 35.8 & 39.1 & 1.3 \\
Random Walk$^*$          & 34.7 & 19.7 & 50.8 & 18.9 & 69.4 & 39.5 & 42.9 & 39.4 & 16.0 & 38.5 & 39.0 & 1.6 & 33.6 & 27.3 & 1.8 \\
Exhaustive Search$^*$    & 33.3 & 28.3 & 62.1 & 33.3 & 70.3 & 63.7 & 66.7 & 61.7 & 22.7 & 36.5 & 51.5 & 3.0 & 38.9 & 70.4 & 2.2 \\
\midrule
\multicolumn{16}{l}{\textit{Active Setting (Multi-turn Spatial Exploration)}} \\
\midrule
Gemini-3.5-flash      & \textbf{59.9} & \textbf{44.8} & \textbf{56.3} & 31.9 & 63.1 & \textbf{68.2} & \textbf{72.2} & \textbf{63.0} & \textbf{32.2} & 28.9 & \textbf{56.8} & 2.5 & \textbf{40.6} & \textbf{80.5} & 9.4 \\
Gemini-3.1-flash-lite & 29.8 & 21.3 & 47.2 & 14.4 & 68.5 & 40.3 & 44.6 & 38.9 & 18.7 & 37.5 & 38.0 & 1.5 & 31.5 & 24.8 & 1.5 \\
Gemini-2.5-flash      & 41.3 & 31.5 & 48.2 & 12.2 & 69.4 & 57.3 & 52.9 & 51.1 & 28.0 & 34.4 & 45.4 & 1.9 & 35.3 & 41.2 & 8.3 \\
Gemini-2.5-flash-lite & 7.1 & 8.7 & 16.9 & 6.7 & 55.0 & 47.6 & 41.3 & 35.0 & 28.0 & 33.3 & 28.1 & 1.6 & 24.5 & 20.6 & 3.1 \\
GPT-5.2-2025-12-11    & 40.4 & 22.0 & 49.2 & \textbf{32.2} & 67.6 & 54.0 & 59.0 & 55.6 & 20.0 & 29.2 & 46.4 & 1.7 & 36.5 & 58.1 & 1.2 \\
GPT-5-mini-2025-08-07 & 34.2 & 23.6 & 50.3 & 10.0 & 59.5 & 50.8 & 49.7 & 46.7 & 28.0 & 34.4 & 41.4 & 1.4 & 34.0 & 29.4 & 3.0 \\
GPT-5-nano-2025-08-07 & 24.9 & 14.2 & 48.7 & 16.7 & 65.8 & 39.5 & 44.6 & 48.3 & 18.7 & 38.5 & 38.0 & 1.3 & 32.0 & 17.6 & 3.6 \\
Qwen2.5-VL-3B-Instruct       & 17.8 & 7.5 & 34.9 & 14.3 & 61.3 & 41.9 & 36.2 & 36.7 & 23.0 & 21.9 & 30.4 & 1.0 & 27.5 & 0.2 & $<$1.0 \\
LLaVA-OneVision2-8B           & 24.0 & 10.2 & 32.3 & 10.0 & 64.9 & 41.9 & 47.8 & 43.9 & 29.3 & \textbf{42.7} & 36.1 & 1.0 & 32.7 & 0.0 & $<$1.0 \\
Gemma-3-4B-it                & 25.3 & 17.7 & 31.8 & 26.8 & 32.8 & 52.4 & 32.7 & 40.0 & 11.5 & 33.3 & 31.0 & 1.1 & 28.0 & 3.3 & $<$1.0 \\
Qwen3-VL-8B-Instruct         & 9.3 & 9.4 & 27.2 & 11.1 & 58.6 & 37.1 & 44.6 & 44.4 & 20.0 & 31.2 & 30.7 & 1.6 & 27.1 & 15.6 & $<$1.0 \\
Qwen3-VL-8B-Thinking         & 12.0 & 11.8 & 19.5 & 8.9 & \textbf{72.1} & 44.4 & 42.3 & 30.6 & 18.7 & 34.4 & 29.8 & 1.1 & 26.7 & 5.8 & $>$20.0 \\
\midrule
\multicolumn{16}{l}{\textit{Passive Setting (Single-turn Static Observation)}} \\
\midrule
Gemini-3.5-flash      & \textbf{61.4} & \textbf{43.4} & \textbf{73.6} & \textbf{43.9} & \textbf{66.0} & \textbf{78.7} & 82.7 & 76.6 & \textbf{54.8} & \textbf{44.4} & \textbf{69.2} & - & - & - & 2.8 \\
Gemini-3.1-flash-lite & 59.0 & 31.6 & 68.9 & 26.8 & 58.5 & 70.2 & 82.7 & 70.8 & 32.3 & 27.8 & 62.6 & - & - & - & $<$1.0 \\
Gemini-2.5-flash      & 47.0 & 27.6 & 65.1 & 19.5 & 60.4 & 61.7 & 75.5 & 73.4 & 25.8 & 36.1 & 58.5 & - & - & - & 1.9 \\
Gemini-2.5-flash-lite & 36.1 & 31.6 & 44.3 & 12.2 & 58.5 & 46.8 & 55.6 & 54.5 & 19.4 & 33.3 & 45.1 & - & - & - & $<$1.0 \\
GPT-5.2-2025-12-11    & 60.2 & 40.8 & 67.9 & 39.0 & 47.2 & 61.7 & 79.6 & \textbf{77.3} & 35.5 & 27.8 & 63.0 & - & - & - & $<$1.0 \\
GPT-5-mini-2025-08-07 & \textbf{61.4} & 42.1 & 65.1 & 34.1 & 47.2 & 67.0 & \textbf{85.7} & \textbf{77.3} & 25.8 & 36.1 & 64.6 & - & - & - & 2.4 \\
GPT-5-nano-2025-08-07 & 44.6 & 35.5 & 62.3 & 24.4 & 49.1 & 62.8 & 67.3 & 69.5 & 16.1 & 36.1 & 55.4 & - & - & - & 1.5 \\
Qwen2.5-VL-3B-Instruct       & 49.4 & 28.2 & 51.9 & 22.6 & 62.1 & 55.3 & 65.3 & 58.4 & 21.6 & 33.3 & 50.9 & - & - & - & $<$1.0 \\
LLaVA-OneVision-2-8B           & 56.6 & 31.6 & 62.3 & 26.8 & 52.8 & 54.3 & 80.1 & 71.4 & 25.8 & 27.8 & 58.9 & - & - & - & $<$1.0 \\
Gemma-3-4B-it                & 33.7 & 25.9 & 37.7 & 26.4 & 25.8 & 54.3 & 49.0 & 51.9 & 24.3 & 25.0 & 40.2 & - & - & - & $<$1.0 \\
Qwen3-VL-8B-Instruct         & 53.0 & 27.6 & 57.5 & 19.5 & 52.8 & 54.3 & 74.0 & 68.2 & 25.8 & 36.1 & 55.6 & - & - & - & $<$1.0 \\
Qwen3-VL-8B-Thinking         & 55.4 & 30.3 & 61.3 & 24.4 & 62.3 & 62.8 & 73.5 & 67.5 & 25.8 & 27.8 & 57.7 & - & - & - & $<$1.0 \\
\bottomrule
\end{tabular}%
}
\end{table*}

\subsection{Benchmark Statistics}
The finalized SATE-Bench comprises 2,525 high-quality, multi-turn QA pairs spanning 10 human-centric tasks. Task distributions and text statistics are detailed in Tables~\ref{tab:dataset_statistics} and~\ref{tab:text_statistics}. The detailed benchmark license, release plan, and more quantitative results are provided in the Appendix~\ref{app:sate_details}.

\subsection{Evaluation Metrics}

\textbf{Accuracy:} The exact-match percentage between the model's final predicted option and the ground-truth answer, evaluated independently of the exploration trajectory. 
To ensure a fair and robust evaluation, we initially apply a rule-based parsing mechanism to extract the model's final choice. If this fails due to irregular formatting, an LLM-based extractor is deployed as a fallback mechanism. 

\noindent\textbf{Exploration Efficiency Reward:} To penalize redundant or excessively long spatial queries, we introduce an exponentially decaying reward metric. For a given task category comprising $N$ questions, the reward is calculated as the average of the step-penalized scores across all questions. Specifically, if a model takes $s_i$ steps to answer the $i$-th question, the task-level exploration reward $R$ is defined as:
\begin{equation}
    R = \frac{1}{N} \sum_{i=1}^{N} \mathbb{I}(c_i) \cdot e^{-\lambda \cdot s_i}
\end{equation}
where $\mathbb{I}(c_i)$ is an indicator function that equals $1$ if the final predicted answer for the $i$-th question is correct and $0$ otherwise, and $\lambda$ is a predefined penalty coefficient. This metric ensures that a correct answer deduced through efficient, deliberate view-exploration yields a higher score than one obtained via random, exhaustive spatial exploration.

\section{Experiments and Main Results}

\subsection{Experimental Setup}

\paragraph{Baselines.} To evaluate MLLM performance and disentangle exploration policies from downstream spatial reasoning, we provide \textbf{9} baselines in total. These include three foundational reference points (\textbf{Random Guess}, \textbf{Blind Text-Only}, and \textbf{Human Baseline}) alongside six evaluation settings applied to Gemini-3.1-flash-lite: \textbf{Oracle-View}, \textbf{Oracle-Trajectory}, \textbf{Always Explore Left}, \textbf{Always Explore Right}, \textbf{Random-Walk}, and \textbf{Exhaustive-Search}. By default, interaction histories are fully visible to models across all applicable evaluations, and the generation temperature is strictly set to $0.0$ to guarantee deterministic and reproducible outputs.
Detailed formulations for all 9 baselines and prompts are provided in the Appendix~\ref{app:exp_details}.


\paragraph{Evaluation Metrics.} For the \textit{Passive} setting (single-turn static observation), we report exact-match \textbf{Average Accuracy (\%)} and total \textbf{Inference Time (hours)}. For the \textit{Active} setting (multi-turn spatial exploration), we provide a more comprehensive assessment to capture interaction dynamics: 1) \textbf{Average Accuracy (\%)}, 2) \textbf{Average Steps} taken per query to reach a final decision, 3) \textbf{Average Reward} ($\lambda=0.1$), which penalizes redundant spatial queries to reflect view-seeking efficiency, 4) \textbf{Active Rate (\%)}, quantifying the frequency at which the model explicitly elects to shift its viewpoint rather than guessing directly from the initial occluded perspective, and 5) \textbf{Inference Time (hours)} for single process under the same hardware settings, while API models are subject to provider latency.

\subsection{Main Results and Discussion}

Based on the comprehensive evaluation across the ten tasks in Table~\ref{tab:main_results}, we identify several insights regarding the active reasoning capabilities of MLLMs:

\noindent\textcolor{teal}{$\blacktriangleright$~\textbf{Finding 1: The Passive-Active Performance Gap.}} There is a severe degradation in accuracy when models transition from passive observation to active spatial exploration. For example, \texttt{Gemini-3.5-flash} achieves a robust 69.2\% accuracy when provided with sufficient static views, but drops to 56.8\% in the Active setting. This discrepancy demonstrates that while MLLMs possess foundational multi-view fusion capabilities, they lack the visual self-awareness required to autonomously sequence exocentric view-selection to resolve severe occlusions.

\noindent\textcolor{teal}{$\blacktriangleright$~\textbf{Finding 2: Premature Spatial Termination.}} Current models suffer from a tendency to halt exploration prematurely. Human evaluators explicitly shift viewpoints in 88.1\% of cases, taking an average of 4.7 steps to achieve 82.7\% accuracy. In contrast, the highest-performing model \texttt{Gemini-3.5-flash} averages only 2.5 steps. This indicates that MLLMs underestimate their own visual uncertainty, falsely concluding they have gathered sufficient geometric evidence when necessary details remain obscured.


\noindent\textcolor{teal}{$\blacktriangleright$~\textbf{Finding 3: Breakdown on Spatial Geometry.}} During multi-turn view-exploration, models handle entity queries reasonably well, such as \texttt{Gemini-3.5-flash} scoring 68.2\% on Object Existence, but fail on tasks requiring precise 3D geometric alignment. Accuracy plummets on Human-Object Interaction (31.9\%) and View-Specific Depth Ordering (32.2\%), highlighting a bottleneck in tracking fine-grained coordinate transformations across dynamic perspective shifts.

\noindent\textcolor{teal}{$\blacktriangleright$~\textbf{Finding 4: The High Inference Cost of Active Exploration.}} Active exploration imposes a severe computational cost. Total inference latency for \texttt{Gemini-3.5-flash} surges to over $236\%$ from 2.8 to 9.4 hours. This highlights that current MLLMs lack the memory-efficient caching required to sustainably process expanding multi-view contexts across continuous perspective exploration.

\begin{table}[tb]
\centering
\caption{Single View Evaluation. In this setting, models are provided strictly with the initial view from the Active set and are evaluated passively (i.e., no spatial exploration is permitted).}
\label{tab:ablation_single_view}
\setlength{\tabcolsep}{3pt}
\resizebox{\columnwidth}{!}{
\begin{tabular}{lclc}
\toprule
\textbf{Model} & \textbf{Avg. Acc (\%)} & \textbf{Model} & \textbf{Avg. Acc (\%)} \\
\midrule
Gemini-3.5-flash      & 37.9 & GPT-5.2-2025-12-11    & 37.3 \\
Gemini-3.1-flash-lite & 37.9 & GPT-5-mini-2025-08-07 & \textbf{39.6} \\
Gemini-2.5-flash      & 37.3 & GPT-5-nano-2025-08-07 & 35.0 \\
\bottomrule
\end{tabular}
}
\end{table}

\noindent\textcolor{teal}{$\blacktriangleright$~\textbf{Finding 5: Open-Source Models Lag in Visual Self-Awareness.}} Despite competitive baseline capabilities, open-source models struggle significantly in the Active setting compared to proprietary systems. While models like \texttt{LLaVA-OneVision2-8B} achieve moderate 58.9\% accuracy in the Passive setting, their Active exploration rates are extraordinarily low, ranging from  0.0\% to 5.8\%. Consequently, their Active accuracies stagnate between 29\% and 36\%, revealing a deficit in spatial proactivity and autonomous multi-turn planning.

\noindent\textcolor{teal}{$\blacktriangleright$~\textbf{Finding 6: Limitations of ``Thinking'' Mechanism in Active Exploration.}} 
In the Passive setting, \texttt{Qwen3-VL-8B-Thinking} leverages extended inference to outperform the \texttt{Instruct} baseline (57.7\% vs. 55.6\%), demonstrating that internal reasoning aids static multi-view fusion. However, this advantage vanishes in the Active setting, where the \texttt{Thinking} model underperforms \texttt{Instruct} (29.8\% vs. 30.7\%). Despite generating extensive chains-of-thought that inflate inference latency ($>20.0$ vs. $<1.0$ hours), it exhibits a much lower active exploration rate (5.8\% vs. 15.6\%). 
This suggests that the thinking mechanism can optimize the internal deduction on provided contexts, but fail to translate this reasoning into spatial proactivity or the recognition of external visual deficits.

\subsection{Single View Evaluation}
To validate that our tasks cannot be solved through single-frame deduction alone, and to show the need for spatial exploration, we restrict models strictly to the initial occluded viewpoint, denying any spatial exploration, as detailed in Table~\ref{tab:ablation_single_view}.

\noindent\textcolor{teal}{$\blacktriangleright$~\textbf{Finding 7: Model Scaling Cannot Overcome Physical Occlusion.}} When trapped in a single ambiguous perspective, scaling model capacity yields no advantage. The compact \texttt{GPT-5-mini} (39.6\%) marginally outperforms the larger models like \texttt{GPT-5.2} (37.3\%). Raw parameter counts cannot reconstruct obscured physical evidence. Without active view-exploration, models can only hallucinate or rely on linguistic priors, validating the difficulty of our benchmark and the necessity of active exploration.


\begin{figure*}[tb]
  \centering
  \includegraphics[width=0.95\textwidth]{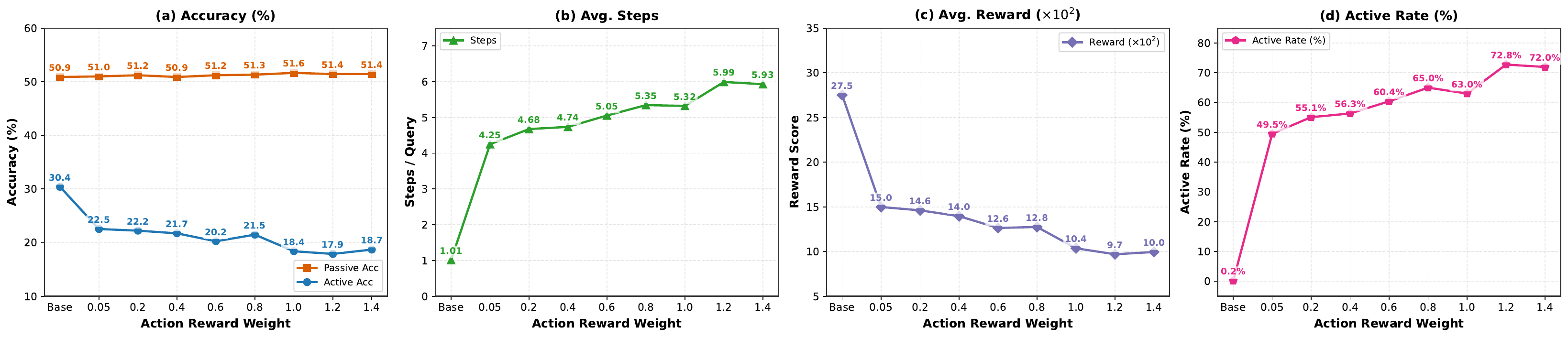}
  \caption{Impact of the action reward weight ($r_3$) during GRPO fine-tuning on Qwen2.5-VL-3B-Instruct. Increasing the action reward successfully incentivizes proactive behavior, evidenced by a sharp rise in (b) average steps and (d) active exploration rate. However, this aggressive exploration exacerbates contextual interference, leading to a steady decline in (a) active accuracy and (c) average reward. In contrast, the model's passive accuracy remains highly stable across all reward configurations.}
  \label{fig:grpo_results}
\end{figure*}

\begin{table}[tb]
\centering
\caption{Evaluation on Expanded Viewpoint Selection. We compare the standard Active setting against an expanded setting where models are provided with broader selection options at each step (allowing shifts of either $\pm 1$ or $\pm 2$ views). The total available views in the 360-degree environment remain consistent (8 views).}
\label{tab:ablation_expanded_views}
\setlength{\tabcolsep}{3pt}
\resizebox{\columnwidth}{!}{
\begin{tabular}{l ccccc}
\toprule
\textbf{Model} & \textbf{Avg.} & \textbf{Avg.} & \textbf{Avg.} & \textbf{Active} & \textbf{Inf. Time} \\
& \textbf{Acc (\%)} & \textbf{Steps} & \textbf{Reward} & \textbf{Rate (\%)} & \textbf{(hours)} \\
\midrule
\multicolumn{6}{l}{\textit{Standard Active ($\pm1$ moves)}} \\
\midrule
Gemini-3.1-flash-lite & 38.0 & 1.5 & 31.5 & 24.8 & 1.5 \\
Gemini-2.5-flash      & 45.4 & 1.9 & 35.3 & 41.2 & 8.3 \\
\midrule
\multicolumn{6}{l}{\textit{Expanded Active ($\pm2$ moves)}} \\
\midrule
Gemini-3.1-flash-lite & 37.9 & 1.4 & 33.1 & 21.0 & 1.1 \\
Gemini-2.5-flash      & 45.7 & 1.9 & 38.0 & 42.9 & 8.7 \\
\bottomrule
\end{tabular}
}
\end{table}

\subsection{Evaluation on Expanded View Selection}
To test if the strictly adjacent ($\pm 1$) view constraint limits performance, we evaluate an expanded setting allowing broader perspective shifts ($\pm 1$ or $\pm 2$), as shown in Table~\ref{tab:ablation_expanded_views}.

\noindent\textcolor{teal}{$\blacktriangleright$~\textbf{Finding 8: Static Accuracy.}}
Despite the capacity to explore perspectives faster, overall accuracy remains virtually unchanged, with \texttt{Gemini-2.5-flash} shifting only marginally from 45.4\% to 45.7\%. This confirms that MLLM spatial reasoning failures stem from a lack of visual self-awareness and failing to recognize uncertainty, rather than restricted querying mechanics.

\noindent\textcolor{teal}{$\blacktriangleright$~\textbf{Finding 9: Marginal Efficiency Gains.}}
While accuracy stagnates, the Average Reward slightly improves from 35.3 to 38.0. When models correctly elect to engage in view-exploration, the expanded options allow them to safely bypass redundant intermediate frames. Consequently, they can reach optimal unoccluded viewpoints more directly, accelerating the resolution of spatial ambiguity and thereby incurring fewer step penalties.

\subsection{Matched-View Evaluation}
To identify the failure, we evaluated a ``Matched-View Passive'' control where the model receives the same views in the identical order as its corresponding Active trajectory.

\noindent\textcolor{teal}{$\blacktriangleright$~\textbf{Finding 10: Active-Passive Gap Stems from Inadequate Exploration.}} 
As shown in Table~\ref{tab:active_vs_passive}, both settings achieve similar accuracy when provided with identical visual evidence. This confirms that the performance drop in the Active setting is primarily caused by insufficient evidence acquisition during exploration, rather than difficulties in processing the multi-turn format itself.

\begin{table}[tb]
\centering
\caption{Performance comparison between Active and Passive settings. Note that ``Passive'' refers to the Matched-View Passive setting. For brevity, model names are abbreviated: ``Gemini-3.1-lite'' refers to Gemini-3.1-flash-lite, and ``GPT-5-mini'' refers to GPT-5-mini-2025-08-07.}
\label{tab:active_vs_passive}
\setlength{\tabcolsep}{3pt}
\resizebox{\columnwidth}{!}{
\begin{tabular}{llcccc}
\toprule
Model & Setting & Avg. Acc (\%) & Avg. Steps & Avg. Reward & Active Rate (\%) \\
\midrule
Gemini-3.1-lite & Active  & 38.0 & 1.5 & 31.5 & 24.8 \\
Gemini-3.1-lite & Passive & 40.3 & -   & -    & -    \\
GPT-5-mini      & Active  & 41.4 & 1.4 & 34.0 & 29.4 \\
GPT-5-mini      & Passive & 41.9 & -   & -    & -    \\
\bottomrule
\end{tabular}
}
\end{table}

\subsection{Long-Context Video Modeling Evaluation}
Recent long-context models could theoretically bypass sequential querying by treating spatial views as a temporal stream. To test this, we formulated all Passive views as a single video sequence.

\noindent\textcolor{teal}{$\blacktriangleright$~\textbf{Finding 11: Video Modality Degrades Reasoning.}}
However, it consistently degrades performance compared to standard multi-image inputs. Specifically, accuracy drops from 62.6\% to 57.6\% for \texttt{Gemini-3.1-flash-lite}, and from 64.6\% to 63.7\% for \texttt{GPT-5-mini}. 
This suggests that current video encoding is less effective at integrating discontinuous cross-camera views that lack natural temporal continuity.
Explicitly framing discrete viewpoints as distinct images remains a more effective strategy.

\subsection{Direct Measurement of Visual Self-Awareness}
To directly measure visual self-awareness, we prompt models to report a continuous confidence score (0--100\%) alongside their action decisions at each exploration step.

\noindent\textcolor{teal}{$\blacktriangleright$~\textbf{Finding 12: Severe Overconfidence Drives Premature Stopping.}} 
As illustrated in Table~\ref{tab:confidence_analysis}, models exhibit severe overconfidence. Despite halting prematurely (averaging $\sim$1.4 steps) and achieving low overall accuracy, they report extremely high terminal confidence scores. This explicitly demonstrates that premature stopping is driven by a false sense of certainty rather than arbitrary policy limits.

\begin{table}[tb]
\centering
\caption{Analysis of exploration behavior and terminal confidence, from a separate run with the confidence-elicitation prompt. Model names are abbreviated for brevity: ``Gemini-3.1-lite'' refers to Gemini-3.1-flash-lite, and ``GPT-5-mini'' refers to GPT-5-mini-2025-08-07.}
\label{tab:confidence_analysis}
\setlength{\tabcolsep}{3pt}
\resizebox{\columnwidth}{!}{
\begin{tabular}{l ccccccc}
\toprule
Model & Avg.     & Avg.  & Avg.   & Active   & Last 3rd & Penult.  & Final \\
      & Acc (\%) & Steps & Reward & Rate (\%)& Conf (\%)& Conf (\%)& Conf (\%) \\
\midrule
Gemini-3.1-lite & 37.3 & 1.4 & 33.0 & 16.2 & 82.1 & 85.2 & 95.1 \\
GPT-5-mini      & 41.4 & 1.4 & 36.4 & 26.3 & 73.2 & 78.5 & 84.7 \\
\bottomrule
\end{tabular}
}
\end{table}

\subsection{Error Analysis}
Our error analysis reveals distinct failure modes across different models. Closed-source models primarily fail due to multi-view fusion bottlenecks and visual hallucinations across accumulated frames. Conversely, open-source models suffer from a fundamental lack of proactivity. Reasoning-focused variants frequently fall into infinite internal deduction loops, exhausting token limits without executing concrete spatial actions. Meanwhile, standard instruction-tuned variants prematurely guess final answers despite severe occlusion. Given this pervasive deficit in autonomous exploration, a critical question arises: can active exploration be elicited through reinforcement learning?

\section{Can MLLMs Learn Active Exploration via Reinforcement Learning?}
Inspired by prior work~\cite{li2026groundingme, de2026proactivebench} and our Findings 5 and 6, we investigate whether MLLMs can acquire active exploration capabilities through Group Relative Policy Optimization (GRPO)~\cite{shao2024deepseekmath}. We synthesize 19.3K training samples from M3GYM and Ego-Exo4D using \texttt{Gemini-3.5-flash}, and fine-tune \texttt{Qwen2.5-VL-3B-Instruct} on a balanced 1:1 mixture of active and passive tasks. To reduce the computational complexity, we simplify the training as a single-turn decision process: the model must either directly answer the question (passive) or select a movement action to explore further (active). The GRPO reward function incorporates an accuracy reward ($r_1=1.0$), a format reward ($r_2=0.5$), and an action reward ($r_3 \in [0.05, 1.4]$). Results are shown in Figure~\ref{fig:grpo_results}. Detailed training data and hyperparameters are provided in the Appendix~\ref{app:train_details}.

\noindent\textcolor{teal}{$\blacktriangleright$~\textbf{Finding 13: GRPO Boosts Active Exploration.}}
Before tuning, the base \texttt{Qwen2.5-VL-3B} exhibits severe ``visual blindness.'' In the Active setting, it yields a mere 0.2\% active exploration rate and averages only 1.0 step per query, preferring to guess blindly rather than navigate away from occluded viewpoints. Post-tuning, the model's willingness to explore surges dramatically; for instance, the model with 1.2 action reward weight achieves a 72.8\% active rate and averages 6.0 steps. This demonstrates that RL effectively incentivizes the behavioral logic of switching views when initial information is insufficient.


\noindent\textcolor{teal}{$\blacktriangleright$~\textbf{Finding 14: Passive Multi-View Perception Remains Stable.}}
When evaluated in the Passive setting, where static, complete multi-view evidence is provided simultaneously, both the Base and GRPO-tuned models maintain highly consistent accuracies ($\sim$51.0\%). This indicates that GRPO tuning does not degrade the model's core multi-image feature fusion capabilities. It further validates the inherent Active--Passive performance gap, confirming that models struggle specifically with the dynamic planning and sequential reasoning required during active exploration.

\noindent\textcolor{teal}{$\blacktriangleright$~\textbf{Finding 15: The Exploration-Accuracy Trade-off.}}
Adjusting the action reward parameter reveals a strict trade-off in policy learning. While higher action rewards incentivize more exploration steps and improve the active rate, they concurrently induce an over-exploration penalty, leading to lower overall accuracy and degraded final rewards.

\section{Conclusion}
We introduce \textbf{SATE-Bench} to evaluate visual self-awareness and active view-exploration in MLLMs. Our evaluations show that models underestimate visual uncertainty under occlusion, stopping prematurely with high confidence. While GRPO incentivizes active view-seeking, over-exploration and multi-view fusion bottlenecks lower accuracy. Ultimately, scaling passive perception cannot yield active spatial reasoning.


\noindent\textbf{Acknowledgements.} This research is funded in part by ARC-Discovery grant (DP220100800, DE230100477, FT260100691 to XY), the Advance Queensland Industry Research Projects (AQIRP), and Follow ME AI Pty LTD. 
This work is also supported by resources provided by The University of Queensland Research Computing Centre’s Bunya supercomputer (https://dx.doi.org/10.48610/wf6c-qy55), with funding from The University of Queensland, Brisbane, Australia.
We thank all anonymous reviewers and ACs for their constructive suggestions.

%% file: sec/X_suppl.tex
\clearpage
\maketitlesupplementary


\startcontents[appendix]

\printcontents[appendix]{l}{1}{\section*{Table of Contents}}
\clearpage

\renewcommand{\thesection}{\Alph{section}}
\setcounter{section}{0}

%
%
%

\section{Details of SATE-Bench}
\label{app:sate_details}

\subsection{More Qualitative Examples}
\begin{figure*}[tb]
  \centering
  \includegraphics[width=\textwidth]{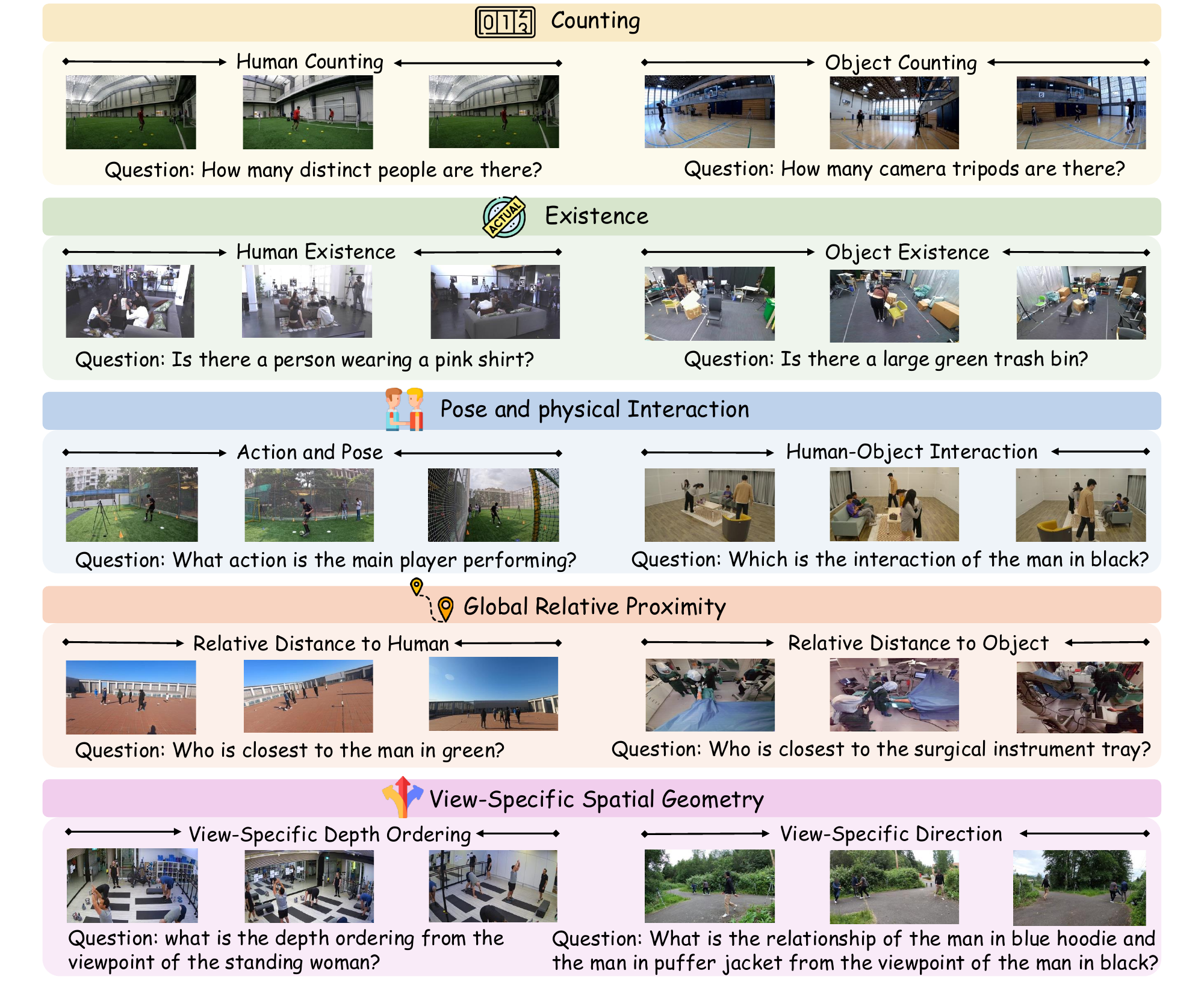}
  \caption{Taxonomy and qualitative examples of SATE-Bench tasks. Our benchmark evaluates multi-view spatial reasoning across five core cognitive dimensions: Counting, Existence, Pose and Physical Interaction, Global Relative Proximity, and View-Specific Spatial Geometry. For each sub-task, we illustrate a representative spatial query alongside a subset of the available exocentric observations required to resolve the scene's geometric ambiguity.
  }
  \label{fig:examples_all}
\end{figure*}

We provide more qualitative examples in Figure~\ref{fig:examples_all}.

\subsection{Details of Source Datasets}
\label{sec:source_dataset_details}

SATE-Bench is constructed from ten high-quality, multi-view, human-centered datasets designed to capture complex real-world spatial interaction scenarios, including severe occlusion, dynamic motion, and multi-person collaboration. These datasets complement each other in terms of scene structure, interaction types, and camera configurations, ensuring broad coverage of human-centered spatial reasoning scenarios.

\paragraph{Core4D~\cite{liu2025core4d}.}
Core4D focuses on human-object-human collaborative interaction in unconstrained 3D environments. It captures scenes where multiple individuals simultaneously manipulate shared objects, often within tightly coupled spatial configurations. By explicitly capturing the intricate dynamics of object handover and joint manipulation, the severe mutual occlusion and overlapping interactions make Core4D particularly well-suited for evaluating cross-view reasoning, object association, and relative spatial localization capabilities in densely populated, shared workspaces.

\paragraph{M3GYM~\cite{xu2025m3gym}.}
M3GYM is a large-scale multi-view motion dataset that captures high-intensity human activity in gym environments. It includes complex body movements such as squats, weightlifting, and stretching exercises, which naturally induce severe self-occlusion and non-rigid pose deformations. The highly synchronized camera rings capture these rapid movements from varying elevations, making it ideal for evaluating fine-grained pose understanding, kinematic tracking, and action recognition under persistent occlusion.

\paragraph{Harmony4D~\cite{khirodkar2024harmony4d}.}
Harmony4D provides intensive multi-camera recordings of human-to-human and human-to-object interactions in the wild. Its synchronized, high-fidelity views enable fine-grained spatial alignment across cameras. By capturing close-proximity activities like wrestling and hugging, it is particularly useful for evaluating close-range physical contact reasoning, depth ordering, and multi-view geometric consistency.

\paragraph{HOI-M3~\cite{zhang2024hoi}.}
HOI-M3 focuses on human-object interaction modeling in multi-view environments, emphasizing the fine-grained contact relationships between human body parts and manipulated objects. By incorporating detailed contextual environments, this dataset presents significant challenges due to uneven object visibility and ambiguous interactivity across different viewpoints, demanding robust 3D geometric projections from observing models.

\paragraph{Ego-Exo4D~\cite{grauman2024ego}.}
Ego-Exo4D is a massive-scale dataset containing paired egocentric (wearable) and other-centric (static exocentric) views of complex skilled human activities. By capturing synchronized activities such as cooking, sports, and mechanics, it enables reasoning across vastly different perspectives, effectively bridging the gap between first-person perception and third-person spatial reconstruction.

\paragraph{EgoHumans~\cite{khirodkar2023ego}.}
EgoHumans captures a wide variety of unscripted human activities in real-world environments from multiple simultaneous perspectives. It emphasizes natural movement and cluttered backgrounds across diverse indoor and outdoor settings. The introduction of varied lighting conditions and scale changes makes it highly suitable for evaluating model robustness under real-world occlusion and dynamic perspective shifts.

\paragraph{MvMHAT~\cite{gan2021mvmhat}.}
MvMHAT is a multi-view dataset designed to understand human movement and multi-person tracking under conditions of severe occlusion. It provides synchronized camera setups that capture subjects' daily activities and interactions in 360-degree omnidirectional footage. The strategic placement of cameras challenges models to maintain spatial awareness of individuals as they transition through heavily occluded blind spots.

\paragraph{Replay~\cite{shapovalov2023replay}.}
Replay offers multi-view recording capabilities with highly controllable camera positions, enabling the capture of both structured and unstructured social activities. Its high-fidelity recording of casual interactions provides rich geometric cues, making it widely used to evaluate temporal consistency, multi-view feature fusion, and robust cross-view correspondence in dynamic scenes.

\paragraph{4D-OR~\cite{ozsoy20224d}.}
4D-OR focuses on the clinical operating room environment, using a dense multi-camera setup to capture the holistic surgical process. The clustering of medical staff around the surgical bed introduces extreme, sustained mutual occlusion caused by personnel and equipment. This makes it highly valuable for detailed spatial reasoning, depth estimation, and entity existence verification in critical medical settings.

\paragraph{MM-OR~\cite{ozsoy2025mm}.}
MM-OR expands upon multimodal operating room data by providing richer step-by-step procedure annotations and enhanced multi-view synchronization capabilities. It serves as a rigorous benchmark designed to aid in understanding complex surgical workflows, tool tracking, and multi-person coordination under conditions of severely limited visual accessibility.

\subsection{Detailed Task Definitions}
\label{sec:task_definition_details}
To rigorously evaluate the capacity of Multimodal Large Language Models (MLLMs) for spatial reasoning and active exocentric view-exploration, we define ten distinct human-centric tasks distributed across five cognitive dimensions.

\paragraph{Entity Counting (Human Counting \& Object Counting).} These tasks determine the total unique instances across complex scenes. In multi-camera environments, counting evaluates global spatial memory and cross-view re-identification. Models must actively resolve localized occlusions via continuous view-exploration, ensuring they track entities across perspective shifts without double-counting.

\paragraph{Entity Existence (Human Existence \& Object Existence).} These tasks verify the presence of specific visual targets that may be completely obscured in the initial view. This fundamentally tests fine-grained searching and visual self-awareness, assessing whether an MLLM can explicitly recognize its observational blind spots and execute an active exploration policy to uncover geometric evidence, rather than relying on language priors to hallucinate an answer.

\paragraph{Pose and Physical Interaction (Fine-grained Action and Pose \& Human-Object Interaction).} Comprehending dense human activity requires granular geometric alignment. These tasks analyze complex body kinematics and exact physical contact points. Models are challenged to resolve severe mutual occlusions and deceptive 2D projections by aggregating spatial evidence through active exocentric exploration.

\paragraph{Global Relative Proximity (Relative Distance to Human \& Relative Distance to Object).} These queries require identifying the closest entity to a designated 3D reference point. This dimension evaluates the model's multi-view depth estimation and cross-perspective distance reasoning, ensuring relative spatial relationships remain consistent even as the model shifts between discrete vantage points.

\paragraph{View-Conditioned Spatial Geometry (View-Specific Depth Ordering \& View-Specific Direction).} These tasks demand dynamic coordinate frame alignment. Models must rank relative depth from a specific target vantage point or predict 2D directional offsets (e.g., top-left) relative to a designated anchor. This directly evaluates perspective-taking and geometric projection, testing if the model can accurately map and transform spatial relationships outside of its current intrinsic observation point.

\subsection{Pre-processing} 
We synchronize videos using official calibration data and extract 4 to 8 views per scene. These are manually sorted in clockwise order, ensuring uniform $360^\circ$ distribution and full scene coverage. This normalizes the querying interface across diverse datasets, enabling standard left/right view exploration.

\subsection{Distractor Design}
Each 5-option QA pair contains one correct answer, three distractors, and one \textit{Abstention Option} (e.g., target is invisible/unanswerable). To prevent models from exploiting linguistic priors without performing visual exploration, distractors are deterministically generated based on misleading view-specific artifacts and semantic/geometric proximity.

\subsection{Paired QA Analysis}
The Active and Passive settings share the same images, questions, and ground-truth answers, differing only in visual initialization. Specifically, while each Passive item provides a comprehensive set of static views, its paired Active initializes the model with a designated occluded viewpoint. Because every Active--Passive pair evaluates the exact same question, the performance gap in Table~\ref{tab:main_results} provides a direct comparison across settings.

\subsection{Data Leakage Analysis}
To rigorously prevent data leakage, all raw images in SATE-Bench are exclusively sourced from the official test splits of the human-centric multi-view datasets. For datasets lacking predefined splits, we manually construct a held-out subset specifically for evaluation. Crucially, all question-answer pairs and spatial annotations are entirely novel, authored explicitly for this benchmark, and have never been publicly released or included in any prior training corpora. 

Similarly, the data utilized for GRPO fine-tuning is strictly isolated from the evaluation set. GRPO training samples are sourced exclusively from the official training splits of the original datasets. For datasets without explicit splits, the training data is sampled from a disjoint subset that strictly excludes any scenes present in SATE-Bench. Furthermore, all QA pairs used for GRPO are newly synthesized, guaranteeing zero overlap with the benchmark annotations and eliminating the risk of test-set memorization.

\subsection{Human Baseline Evaluation}
\begin{figure*}[tb]
  \centering
  \includegraphics[width=\textwidth]{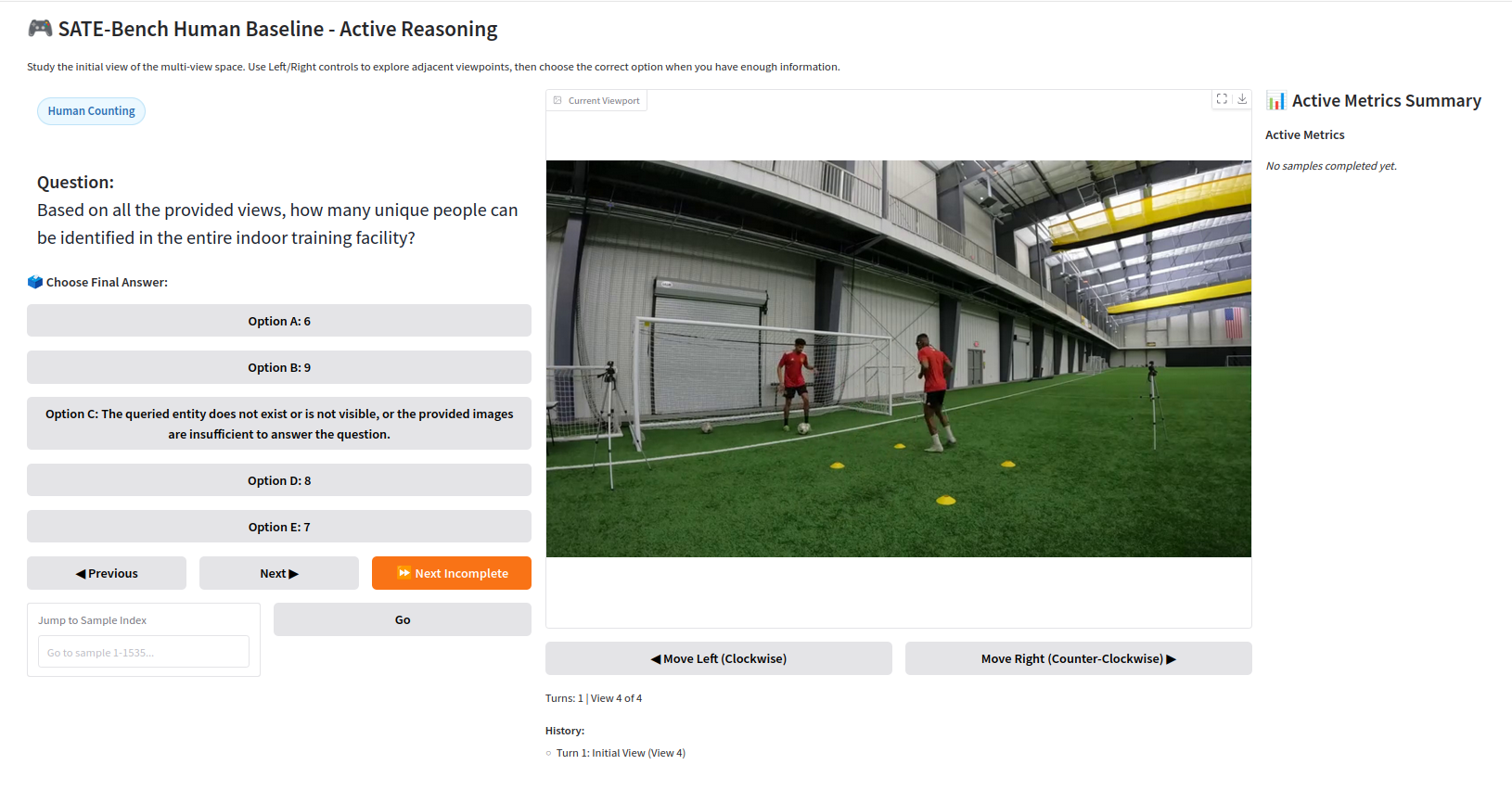}
  \caption{The interactive evaluation interface for the human performance baseline.}
  \label{fig:human_baseline}
\end{figure*}

To establish a rigorous human performance upper bound, we recruit five independent postgraduate evaluators with backgrounds in Computer Science and Biomechanics. Participants complete the tasks without time restrictions via a custom interactive Gradio interface, shown in Figure~\ref{fig:human_baseline}, which allows them to seamlessly execute discrete exocentric view-exploration queries. To mitigate interface familiarity bias, all evaluators undergo a guided 20-question trial phase prior to the formal assessment. Crucially, to guarantee a fair comparison, human participants are subjected to the identical step-wise exploration constraints as the evaluated MLLMs. They receive the exact same system prompts, task instructions, and environmental context, including the total camera count and circular layout, ensuring they navigate the 360-degree scenes under identical restricted visual conditions.

\subsection{License of Source Datasets and Release Plan}
We categorize the source datasets by their access protocols. As detailed in Table \ref{tab:dataset_licenses}, we withhold raw imagery for any dataset that requires a signed user agreement and release only metadata extraction scripts in those cases, ensuring future users must first pass the original creators' ethical vetting processes before using SATE-Bench.

We will fully release and maintain the QA annotations and multi-view images for SATE-Bench and training on GitHub and Hugging Face under a non-commercial license.

\begin{table*}[tb]
\centering
\caption{Overview of source datasets categorized by licensing and access requirements. To ensure compliance with original distribution terms, raw imagery is withheld for datasets requiring signed agreements.}
\label{tab:dataset_licenses}
\resizebox{\textwidth}{!}{%
\begin{tabular}{llcllc}
\toprule
Dataset & License Type & Publicly Available & Access Requirement & Released Artifacts & Re-identification \\
\midrule
M3GYM     & Non-Commercial Academic & Yes & Requires Signed Agreement & Extraction Scripts & No \\
Ego-Exo4D & Ego4D License Agreement & Yes & Requires Signed Agreement & Extraction Scripts & No \\
4D-OR     & MIT                     & Yes & Requires Signed Agreement & Extraction Scripts & No \\
MM-OR     & Apache-2.0              & Yes & Requires Signed Agreement & Extraction Scripts & No \\
Harmony4D & MIT                     & Yes & Direct Download           & Multi-View Frames  & No \\
HOI-M3    & Apache-2.0              & Yes & Direct Download           & Multi-View Frames  & No \\
EgoHumans & Apache-2.0              & Yes & Direct Download           & Multi-View Frames  & No \\
MvMHAT    & Non-Commercial Academic & Yes & Direct Download           & Multi-View Frames  & No \\
Replay    & CC BY-NC 4.0            & Yes & Direct Download           & Multi-View Frames  & No \\
core4D    & CC BY 4.0               & Yes & Direct Download           & Multi-View Frames  & No \\
\bottomrule
\end{tabular}%
}
\end{table*}

\section{Detailed Failure Case Analysis}
\label{sec:supp_failure}

\tcbset{failcase/.style={colback=white, colframe=black!60, boxrule=0.5pt, arc=1mm,
  left=1.5mm, right=1.5mm, top=1mm, bottom=1mm, before skip=0pt, after skip=2mm,
  fonttitle=\small\bfseries, fontupper=\footnotesize}}
\providecommand{\cmark}{\textcolor{green!50!black}{\checkmark}}
\providecommand{\xmark}{\textcolor{red!70!black}{$\times$}}

We first quantify how models fail in each setting in Table~\ref{tab:supp_failure_modes} and then show nine representative cases: five Active cases in Figures~\ref{fig:supp_case_a1} (A1--A3) and~\ref{fig:supp_case_a4} (A4--A5), and four Passive cases in Figures~\ref{fig:supp_case_p1} (P1--P2) and~\ref{fig:supp_case_p3} (P3--P4).
Each case shows the camera views, the trajectories or predictions of models, an excerpt of one model's reasoning summary, and an analysis of where the error arises. In the figures, the orange frame marks the start view, blue frames mark the views visited by the model (Active, numbered in visit order) or given as input (Passive), and grey frames mark views that were not visited or not given. Green boxes mark the decisive evidence, and the yellow box marks the reference person.

\begin{table}[!htb]
\centering
\caption{Distribution of active failure cases (\%). Acc.: Active accuracy; T1: answered (Ans.) or abstained (Abs.) at the start view; Part.: answered after a partial exploration; All: answered after observing all views; None: no valid answer.}
\label{tab:supp_failure_modes}
\resizebox{\columnwidth}{!}{%
\setlength{\tabcolsep}{3pt}
\begin{tabular}{lrrrrrr}
\toprule
 & & \multicolumn{2}{c}{No move} & \multicolumn{3}{c}{After moving} \\
\cmidrule(lr){3-4}\cmidrule(lr){5-7}
Model & Acc. & T1-Ans. & T1-Abs. & Part. & All & None \\
\midrule
Gemini-3.5-flash & 56.8 & 19.0 & 2.0 & 65.1 & 13.5 & 0.4 \\
Gemini-2.5-flash & 45.4 & 45.9 & 15.8 & 28.4 & 9.2 & 0.7 \\
Gemini-3.1-flash-lite & 38.0 & 50.6 & 25.1 & 17.0 & 6.9 & 0.3 \\
Gemini-2.5-flash-lite & 28.1 & 26.2 & 57.2 & 9.9 & 0.5 & 6.3 \\
GPT-5.2 & 46.4 & 34.5 & 9.5 & 54.9 & 0.9 & 0.2 \\
GPT-5-mini & 41.4 & 67.0 & 4.0 & 27.4 & 1.3 & 0.3 \\
GPT-5-nano & 38.0 & 69.7 & 10.5 & 19.1 & 0.4 & 0.2 \\
Qwen3-VL-8B-Instruct & 30.7 & 45.9 & 36.4 & 7.8 & 3.2 & 6.8 \\
Qwen3-VL-8B-Thinking & 29.8 & 48.5 & 45.5 & 4.5 & 0.6 & 0.8 \\
LLaVA-OneVision-2-8B & 36.1 & 69.5 & 30.5 & 0.0 & 0.0 & 0.0 \\
\bottomrule
\end{tabular}}
\end{table}

\subsection{Active Failure Modes}
\begin{figure*}[p]
\centering
\begin{tcolorbox}[failcase, title={Case A1 (Active): premature stop driven by a dataset prior}]
\includegraphics[width=\linewidth]{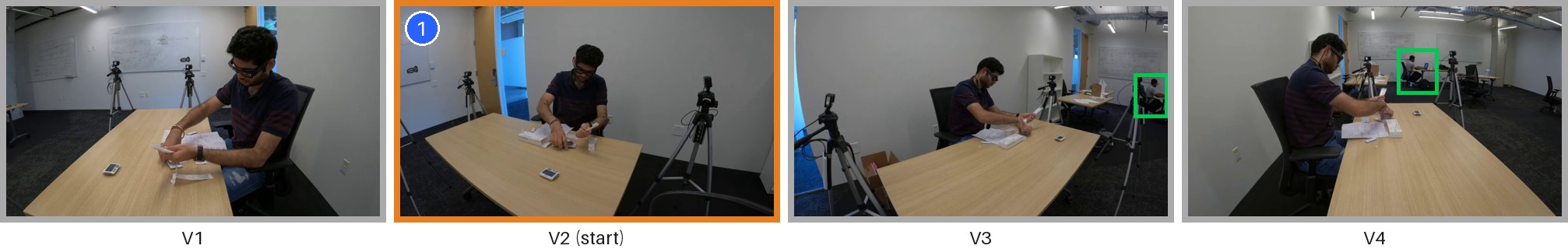}\par\smallskip
{\color{black!60}Ego-Exo4D $\cdot$ Human Counting $\cdot$ 4 views $\cdot$ start V2}\par
\textbf{Q:} How many unique human individuals are present in the entire scene across all given views?
\textbf{Options:} (A)~Abstention \ (B)~1 \ (C)~3 \ (D)~4 \ (E)~2. \ \textbf{GT:} (E).\par
\textbf{Trajectories:} 6 models incl.\ Gemini-3.5-flash: V2 $\Rightarrow$ B \xmark; \ GPT-5.2, GPT-5-mini: V2$\to$V3 $\Rightarrow$ E \cmark; \ Gemini-2.5-flash: V2$\to$V3$\to$V4$\to$V1 $\Rightarrow$ E \cmark.\par
\textbf{Gemini-3.5-flash (V2):} \textit{``Typically, these scenarios focus on a single participant [\dots] exploring other views seems unnecessary.''}\par
\textbf{Analysis:} A second person, seated at a desk in the far corner, is visible only from V3 and V4. The model replaces visual verification with a prior about the source dataset; a single move to V3 would reveal the second person.
\end{tcolorbox}

\begin{tcolorbox}[failcase, title={Case A2 (Active): absence of evidence taken as evidence of absence}]
\includegraphics[width=\linewidth]{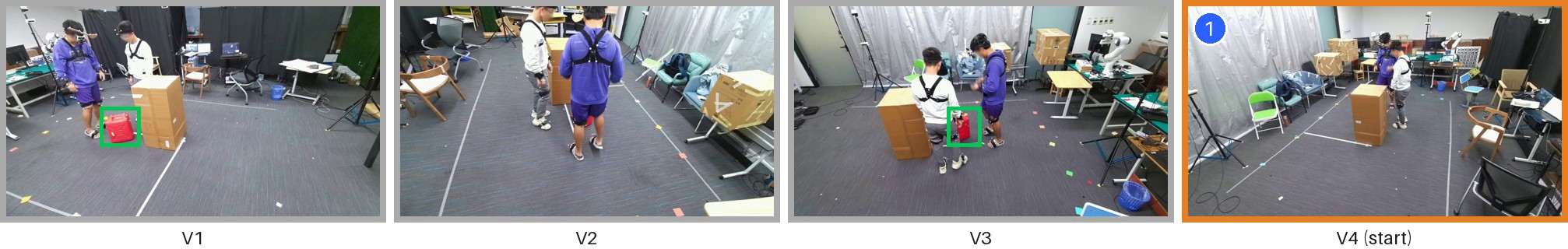}\par\smallskip
{\color{black!60}Core4D $\cdot$ Object Existence $\cdot$ 4 views $\cdot$ start V4}\par
\textbf{Q:} Is there a red plastic fuel container located on the floor between the two participants and the central cardboard boxes?
\textbf{Options:} (A)~No \ (B)~Yes. \ \textbf{GT:} (B).\par
\textbf{Trajectories:} 7 models incl.\ Gemini-3.5-flash: V4 $\Rightarrow$ A \xmark; \ GPT-5.2, GPT-5-nano: V4$\to$V1 $\Rightarrow$ B \cmark; \ Gemini-2.5-flash-lite: V4$\to$V3 $\Rightarrow$ B \cmark.\par
\textbf{Gemini-3.5-flash (V4):} \textit{``I am sure, if it was there, I would see it. [\dots] that space is completely visible and empty.''}\par
\textbf{Analysis:} From V4 the container is hidden behind the tall box and the participants; in V1 and V3 it is clearly visible. The model wrongly judges the region to be fully observed, which is exactly the self-assessment the Active setting probes.
\end{tcolorbox}

\begin{tcolorbox}[failcase, title={Case A3 (Active): abstaining instead of exploring}]
\includegraphics[width=\linewidth]{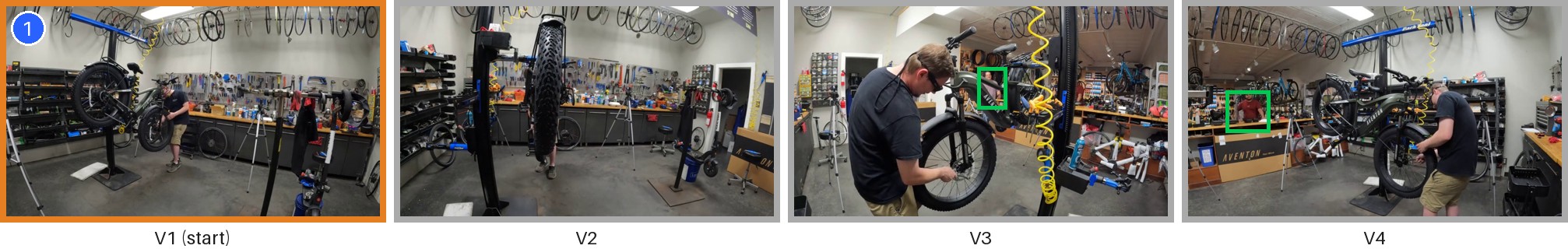}\par\smallskip
{\color{black!60}Ego-Exo4D $\cdot$ Relative Distance to Object $\cdot$ 4 views $\cdot$ start V1}\par
\textbf{Q:} Which person is standing closest to the man wearing the red shirt behind the wooden counter?
\textbf{Options:} (A)~The mechanic in the black shirt at the repair stand \ (B)~The customer entering through the far door \ (C)~Abstention \ (D)~The camera operator in the center of the room \ (E)~The person in the lavender shirt near the background racks. \ \textbf{GT:} (E).\par
\textbf{Trajectories:} 8 models incl.\ Gemini-3.5-flash: V1 $\Rightarrow$ C \xmark; \ Gemini-2.5-flash: V1$\to$V2$\to$V3 $\Rightarrow$ E \cmark; \ GPT-5-nano: V1 $\Rightarrow$ E \cmark\ (unsupported guess).\par
\textbf{Gemini-3.5-flash (V1):} \textit{``If there's no red-shirted guy anywhere, then C is the answer. [\dots] No space for someone to be behind it!''}\par
\textbf{Analysis:} Both persons are outside V1 (red shirt in V4, lavender shirt in V3). The model notices that the referent is missing but declares the question unanswerable instead of exploring, whereas Gemini-2.5-flash explores from the same observation. Turn-1 abstention causes 25--57\% of the errors of the weaker models (Table~\ref{tab:supp_failure_modes}).
\end{tcolorbox}
\caption{Active failure cases A1--A3: the model stops at the start view.}
\label{fig:supp_case_a1}
\end{figure*}

\begin{figure*}[p]
\centering
\begin{tcolorbox}[failcase, title={Case A4 (Active): wrong direction, and evidence not linked across views}]
\includegraphics[width=\linewidth]{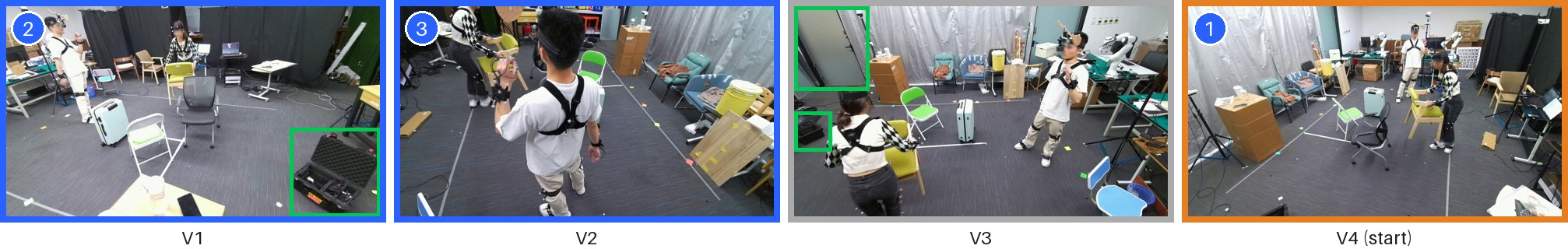}\par\smallskip
{\color{black!60}Core4D $\cdot$ Object Existence $\cdot$ 4 views $\cdot$ start V4}\par
\textbf{Q:} Is there an open black equipment case on the floor near the glass door?
\textbf{Options:} (A)~No \ (B)~Yes. \ \textbf{GT:} (B).\par
\textbf{Trajectories:} Gemini-3.5-flash: V4$\to$V1$\to$V2 $\Rightarrow$ A \xmark; \ Gemini-2.5-flash, GPT-5.2: V4$\to$V3 $\Rightarrow$ B \cmark; \ other 7 models: V4 $\Rightarrow$ A \xmark.\par
\textbf{Gemini-3.5-flash:} \textit{(V1) ``there's this black equipment case -- it's open [\dots] Where is the glass door?'' \ (V2) ``No black case. Just cables.''}\par
\textbf{Analysis:} The case is visible in V1 but the door is not; both appear together only in V3, one step in the opposite direction. The model finds the object, keeps moving away from the view that relates it to the landmark, and forgets the earlier sighting at V2. Partial exploration accounts for 65\% of Gemini-3.5-flash's and 55\% of GPT-5.2's errors.
\end{tcolorbox}

\begin{tcolorbox}[failcase, title={Case A5 (Active): same evidence, different conclusions}]
{\centering\includegraphics[width=\linewidth,height=0.2\textheight,keepaspectratio]{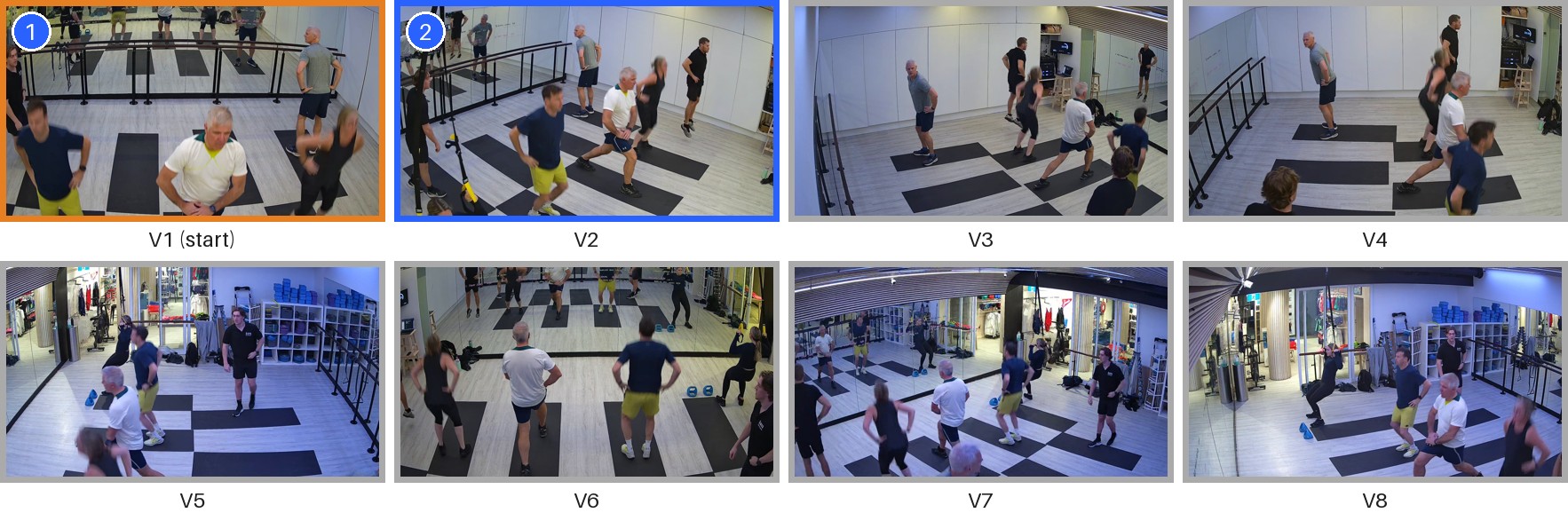}\par}\smallskip
{\color{black!60}M3GYM $\cdot$ Human Counting $\cdot$ 8 views $\cdot$ start V1}\par
\textbf{Q:} How many unique real people are present in the gym space from all given views, including those partially visible or in reflections?
\textbf{Options:} (A)~8 \ (B)~5 \ (C)~6 \ (D)~Abstention \ (E)~7. \ \textbf{GT:} (E).\par
\textbf{Trajectories:} Gemini-3.5-flash: V1$\to$V2 $\Rightarrow$ A \xmark; \ Gemini-2.5-flash, GPT-5.2, GPT-5-mini: V1$\to$V2 $\Rightarrow$ E \cmark; \ GPT-5-nano: V1$\to$V2$\to$V3 $\Rightarrow$ E \cmark; \ Gemini-3.1-flash-lite, Qwen3-VL-8B-Thinking: V1 $\Rightarrow$ A \xmark; \ LLaVA-OneVision-2-8B: V1 $\Rightarrow$ D \xmark; \ Qwen3-VL-8B-Instruct, Gemini-2.5-flash-lite: step limit, no answer \xmark.\par
\textbf{Gemini-3.5-flash (V2):} \textit{``this `new' person (in the reflection) has black shorts [\dots] I will now identify as Person 8.''}\par
\textbf{Gemini-2.5-flash (V2):} \textit{``the two `reflection-only' figures from View 1 are directly visible in View 2, meaning those are not unique individuals.''}\par
\textbf{Analysis:} Both Gemini models observe the same two views. The mirror wall duplicates most participants; Gemini-3.5-flash re-identifies people by a single clothing cue and counts a reflection as an extra person, while Gemini-2.5-flash matches it to a directly visible person. Here the bottleneck is cross-view integration, not exploration.
\end{tcolorbox}
\caption{Active failure cases A4--A5: errors after exploring.}
\label{fig:supp_case_a4}
\end{figure*}

We group the wrong Active episodes by when the model stops: it answers or abstains at the start view, answers after visiting only some of the views, answers wrongly after visiting all views, or never gives a valid answer. For seven of the ten models in Table~\ref{tab:supp_failure_modes}, 71--100\% of the errors are made at the start view without a single move (Figure~\ref{fig:supp_case_a1}), and LLaVA-OneVision-2-8B never moves at all. Models that explore more often fail in a different way. For Gemini-3.5-flash and GPT-5.2, 65\% and 55\% of the errors, respectively, occur after a partial exploration that stops before the decisive view is reached (Case~A4, Figure~\ref{fig:supp_case_a4}). Another 13.5\% of Gemini-3.5-flash's errors are made after it has seen every view, so the difficulty there lies in combining evidence across views rather than in finding it (Case~A5, Figure~\ref{fig:supp_case_a4}).

\subsection{Passive Failure Modes}
\begin{figure*}[p]
\centering
\begin{tcolorbox}[failcase, title={Case P1 (Passive): hallucinated contact}]
\includegraphics[width=\linewidth]{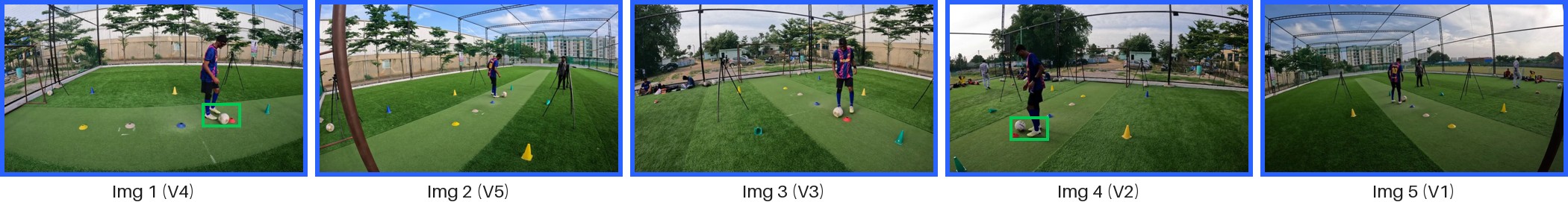}\par\smallskip
{\color{black!60}Ego-Exo4D $\cdot$ Human-Object Interaction $\cdot$ 5 input views}\par
\textbf{Q:} Regarding the interaction between the player's right foot and the soccer ball, which description is most accurate based on the synchronized views?
\textbf{Options:} (A)~Abstention \ (B)~The player's right toes are in firm contact with the top of the ball \ (C)~The player is standing on the ball with the sole of his right foot \ (D)~No contact; the foot is hovering just beside the ball \ (E)~No contact; the foot is hovering just above the ball. \ \textbf{GT:} (D).\par
\textbf{Predictions:} Gemini-3.5-flash: C \xmark; \ Gemini-2.5-flash-lite, GPT-5-mini, GPT-5-nano, Qwen3-VL-8B-Instruct: B \xmark; \ LLaVA-OneVision-2-8B: E \xmark; \ the other 4 models: D \cmark.\par
\textbf{Gemini-3.5-flash:} \textit{``Image 1: Right foot is on the ball. [\dots] Image 5: Right foot is on the ball. [\dots] D \& E) Not likely. There is contact.''}\par
\textbf{Analysis:} The two close-ups (Img 1 and Img 4) show both feet on the turf beside the ball. The model asserts a stereotyped pose first and then ``confirms'' it in every view, including distant ones where the foot covers only a few pixels.
\end{tcolorbox}

\begin{tcolorbox}[failcase, title={Case P2 (Passive): answering in the camera's frame instead of the observer's}]
{\centering\includegraphics[width=\linewidth,height=0.2\textheight,keepaspectratio]{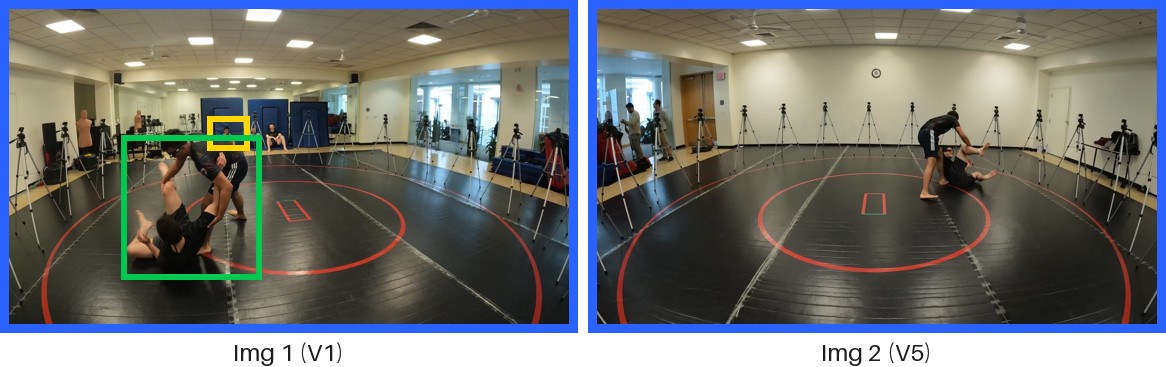}\par}\smallskip
{\color{black!60}Harmony4D $\cdot$ View-Specific Direction $\cdot$ 2 input views}\par
\textbf{Q:} From the perspective of the man sitting in the blue T-shirt, what is the 3D spatial relationship of the standing wrestler relative to the grounded wrestler's torso?
\textbf{Options:} (A)~Behind, right, above \ (B)~Behind, left, below \ (C)~In front, right, below \ (D)~In front, left, above \ (E)~Abstention. \ \textbf{GT:} (D).\par
\textbf{Predictions:} Gemini-2.5-flash, Qwen3-VL-8B-Thinking: A \xmark; \ Gemini-2.5-flash-lite: C \xmark; \ the other 7 models: D \cmark.\par
\textbf{Gemini-2.5-flash:} \textit{``From the observer's perspective, the grounded wrestler's torso is on the left side of the scene, and the standing wrestler is to the right.''}\par
\textbf{Analysis:} The observer (yellow box) faces the camera, so his frame is rotated by about 180$^\circ$ relative to V1. Option A is the correct answer in the camera's frame; for the observer, depth and lateral components flip. The model claims to take the observer's perspective but reads left/right and depth directly off the image.
\end{tcolorbox}
\caption{Passive failure cases P1--P2.}
\label{fig:supp_case_p1}
\end{figure*}

\begin{figure*}[p]
\centering
\begin{tcolorbox}[failcase, title={Case P3 (Passive): accepting a partial attribute match}]
\includegraphics[width=\linewidth]{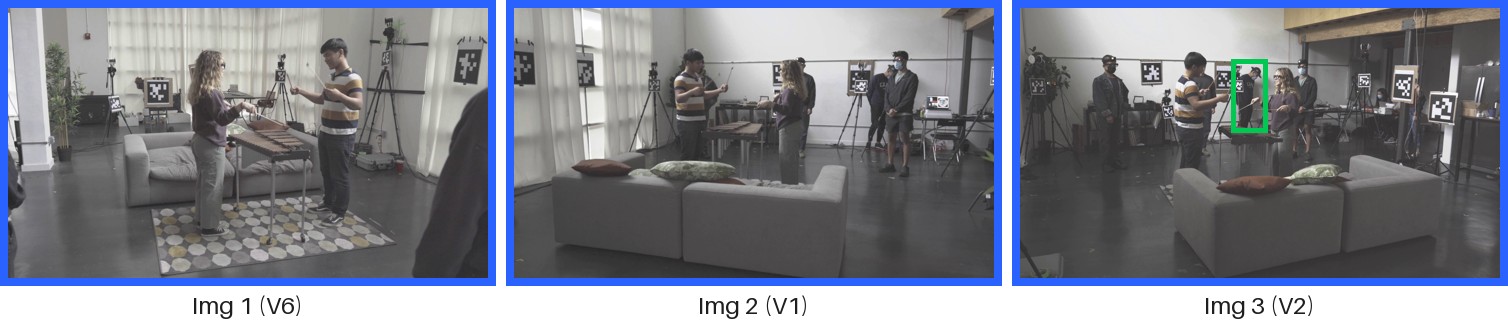}\par\smallskip
{\color{black!60}Replay $\cdot$ Human Existence $\cdot$ 3 input views}\par
\textbf{Q:} Is there a man wearing a grey hooded sweatshirt and a black baseball cap, looking to his right and standing directly behind the two marimba players?
\textbf{Options:} (A)~Yes \ (B)~No. \ \textbf{GT:} (B).\par
\textbf{Predictions:} Gemini-3.5-flash, Gemini-2.5-flash, GPT-5-mini, GPT-5-nano, Qwen3-VL-8B-Instruct, Qwen3-VL-8B-Thinking: A \xmark; \ the other 4 models: B \cmark.\par
\textbf{Gemini-3.5-flash:} \textit{``His hoodie's possibly dark grey but looks like black in Image 2. [\dots] this seems to be the closest match.''}\par
\textbf{Analysis:} The man behind the players (Img 3) wears a black cap but a dark navy sweatshirt; a man in grey without a cap stands next to him. The question mixes attributes of two people. The model notices the mismatch but accepts the closest match instead of rejecting it, the typical form of the 76\% false-positive existence errors.
\end{tcolorbox}

\begin{tcolorbox}[failcase, title={Case P4 (Passive): merging two instances across views}]
\includegraphics[width=\linewidth]{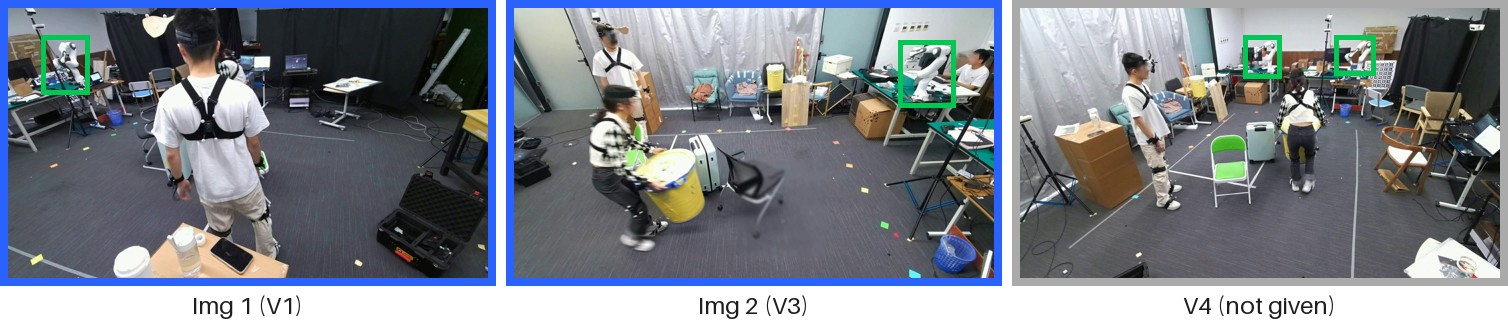}\par\smallskip
{\color{black!60}Core4D $\cdot$ Object Counting $\cdot$ 2 input views}\par
\textbf{Q:} How many white robotic arms are installed on the workbenches in the space?
\textbf{Options:} (A)~1 \ (B)~3 \ (C)~2 \ (D)~Abstention \ (E)~0. \ \textbf{GT:} (C).\par
\textbf{Predictions:} Gemini-3.5-flash, Gemini-2.5-flash, Gemini-3.1-flash-lite, LLaVA-OneVision-2-8B: A \xmark; \ GPT-5-mini: B \xmark; \ the other 5 models: C \cmark.\par
\textbf{Gemini-3.5-flash:} \textit{``The positioning suggests it could be the same workbench. [\dots] Same workbench. Same arm.''}\par
\textbf{Analysis:} The room has two arms on adjacent workbenches; Img 1 and Img 2 each show a different one (V4, not given as input, shows both). The model identifies the real question, i.e., whether the two sightings are the same instance, but decides it from generic context that fits both benches. Such over-merging explains why under-counts are twice as frequent as over-counts.
\end{tcolorbox}
\caption{Passive failure cases P3--P4.}
\label{fig:supp_case_p3}
\end{figure*}

Across all the models evaluated in the Passive setting, three error types are the most common. (i) \emph{Under-counting}: 66\% of the counting errors with a numeric answer are under-counts (422 vs.\ 217), usually because distinct instances seen in different views are counted as one (Case~P4, Figure~\ref{fig:supp_case_p3}). (ii) \emph{False positives in existence questions}: 76\% of the existence errors answer ``Yes'' for an absent target (343 vs.\ 106), usually because the model accepts an entity that matches only part of the description (Case~P3, Figure~\ref{fig:supp_case_p3}). (iii) \emph{Frame-of-reference errors} in view-specific direction questions: of the 48 errors whose options can be split into depth, lateral and vertical components, 36 flip only left and right, and 3 give the answer in the camera's frame instead of the observer's (Case~P2, Figure~\ref{fig:supp_case_p1}). In interaction questions, models also often report physical contact that does not exist (Case~P1, Figure~\ref{fig:supp_case_p1}).

\section{Experiment Details}
\label{app:exp_details}

\subsection{Details of the baselines}
\paragraph{Baselines.} To rigorously ground MLLM performance and disentangle the exploration policy from downstream spatial reasoning, we provide 9 baselines in total. By default, interaction histories are fully visible to models for all evaluations. The baselines include:

\begin{itemize}
    \item \textbf{Random Guess:} Assumes uniform selection among given choices, yielding expected accuracies of 20\% for 5-option formats and 50\% for 2-option formats.
    
    \item \textbf{Blind (Text-Only):} Evaluates Gemini-2.5-flash exclusively on text prompts with all visual input completely withheld. This quantifies the extent of exploitable linguistic priors in the benchmark.
    
    \item \textbf{Human Baseline:} Averages the performance of five independent graduate evaluators (distinct from the annotation team). For the Active setting, these subjects interact with the identical step-wise $360^\circ$ environment, establishing an upper bound for spatial reasoning and active exploration accuracy.
    
    \item \textbf{Oracle-View:} Evaluates performance when the model is directly provided with the single most informative viewpoint containing the unoccluded evidence.
    
    \item \textbf{Oracle-Trajectory:} Supplies the model with the optimal sequence of navigational steps to uncover the hidden evidence, isolating reasoning capabilities from path-planning failures.
    
    \item \textbf{Always Explore Left:} Forces the model to follow a predetermined, rigid sequence of viewpoints by continuously panning to the left. This provides smaller gains compared to active planning but isolates the benefit of a simple, fixed exploration trajectory.
    
    \item \textbf{Always Explore Right:} Forces the model to continuously pan to the right. Together with the left-exploration baseline, this assesses whether simple fixed-order traversal is sufficient and checks for directional bias in the environment.
    
    \item \textbf{Random-Walk Exploration:} The agent traverses the environment using randomly selected navigational steps. This baseline typically offers minimal improvement, highlighting the necessity of deliberate planning over stochastic movement.
    
    \item \textbf{Exhaustive-Search:} Allows the model to sweep all possible viewpoints in the $360^\circ$ environment before answering, testing its capacity for cross-view integration and memory retention over long contexts.
\end{itemize}

Through these variations (the latter six evaluated using Gemini-3.1-flash-lite), we can pinpoint specific bottlenecks. For instance, comparing the active policy against Oracle, fixed-order, and random baselines reveals that premature stopping and ineffective viewpoint selection are major failure modes in the Active setting, though imperfect cross-view integration remains a contributing factor.

\subsection{Full table for Single View Evaluation}
The full results for single-view evaluation are provided in Table~\ref{tab:ablation_single_view_full}.

\begin{table*}[tb]
\centering
\caption{Full Table for Single View Evaluation. In this setting, models are provided strictly with the initial view from the Active set and are evaluated passively (i.e., no spatial exploration is permitted).}
\label{tab:ablation_single_view_full}
\resizebox{\textwidth}{!}{%
\begin{tabular}{l|cccccccccc|c}
\toprule
\textbf{Model} & \textbf{HC} & \textbf{OC} & \textbf{FAP} & \textbf{HOI} & \textbf{HE} & \textbf{OE} & \textbf{RD-H} & \textbf{RD-O} & \textbf{VDO} & \textbf{VSD} & \textbf{Avg.} \\
& & & & & & & & & & & \textbf{Acc (\%)} \\
\midrule
Gemini-3.5-flash      & 25.8 & 16.5 & 39.0 & 15.6 & 65.8 & \textbf{50.0} & 47.4 & 42.8 & \textbf{29.3} & 31.2 & 37.9 \\
Gemini-3.1-flash-lite & 25.3 & 15.0 & \textbf{49.7} & 15.6 & 70.3 & 37.9 & 47.1 & 39.4 & 18.7 & \textbf{38.5} & 37.9 \\
Gemini-2.5-flash      & 24.9 & 15.7 & 40.0 & 12.2 & \textbf{77.5} & \textbf{50.0} & 51.0 & 34.4 & 17.3 & 27.1 & 37.3 \\
Gemini-2.5-flash-lite & 19.1 & 15.0 & 17.4 & 13.3 & 69.4 & 37.1 & 31.7 & 31.7 & 26.7 & 37.5 & 28.9 \\
GPT-5.2-2025-12-11    & 26.7 & 14.2 & 36.9 & \textbf{21.1} & 61.3 & 42.7 & \textbf{54.5} & 38.3 & 17.3 & 32.3 & 37.3 \\
GPT-5-mini-2025-08-07 & \textbf{32.0} & \textbf{17.3} & 45.6 & 6.7 & 62.2 & 46.0 & 50.0 & \textbf{47.8} & \textbf{29.3} & 30.2 & \textbf{39.6} \\
GPT-5-nano-2025-08-07 & 24.4 & 13.4 & 40.0 & 8.9 & 63.1 & 38.7 & 42.9 & 45.6 & 18.7 & 33.3 & 35.0 \\
\bottomrule
\end{tabular}%
}
\end{table*}

\subsection{Full table for Expanded View Selection}
The full results for expanded view selection evaluation are provided in Table~\ref{tab:ablation_expanded_views_full}.

\begin{table*}[tb]
\centering
\caption{Full Table for Expanded Viewpoint Selection. We compare the standard Active setting against an expanded setting where models are provided with broader selection options at each step (allowing shifts of either $\pm 1$ or $\pm 2$ views). The total available views in the 360-degree environment remain consistent (8 views). Task abbreviations follow the main table. Metrics are presented for \texttt{Gemini-3.1-flash-lite} and \texttt{Gemini-2.5-flash} to analyze the impact of increased spatial querying flexibility on exploration efficiency.}
\label{tab:ablation_expanded_views_full}
\resizebox{\textwidth}{!}{%
\begin{tabular}{l|cccccccccc|ccccc}
\toprule
\textbf{Model} & \textbf{HC} & \textbf{OC} & \textbf{FAP} & \textbf{HOI} & \textbf{HE} & \textbf{OE} & \textbf{RD-H} & \textbf{RD-O} & \textbf{VDO} & \textbf{VSD} & \textbf{Avg.} & \textbf{Avg.} & \textbf{Avg.} & \textbf{Active} & \textbf{Inf. Time} \\
& & & & & & & & & & & \textbf{Acc (\%)} & \textbf{Steps} & \textbf{Reward} & \textbf{Rate (\%)} & \textbf{(hours)} \\
\midrule
\multicolumn{16}{l}{\textit{Original Active Setting (Unrestricted 8-View Exploration)}} \\
\midrule
Gemini-3.1-flash-lite & 29.8 & 21.3 & 47.2 & 14.4 & 68.5 & 40.3 & 44.6 & 38.9 & 18.7 & 37.5 & 38.0 & 1.5 & 31.5 & 24.8 & 1.5 \\
Gemini-2.5-flash      & 41.3 & 31.5 & 48.2 & 12.2 & 69.4 & 57.3 & 52.9 & 51.1 & 28.0 & 34.4 & 45.4 & 1.9 & 35.3 & 41.2 & 8.3 \\
\midrule
\multicolumn{16}{l}{\textit{Ablated Active Setting (Restricted to Maximum $\pm2$ Views)}} \\
\midrule
Gemini-3.1-flash-lite & 30.7 & 18.9 & 45.6 & 16.7 & 68.5 & 41.1 & 45.2 & 36.1 & 20.0 & 37.5 & 37.9 & 1.4 & 33.1 & 21.0 & 1.1 \\
Gemini-2.5-flash      & 47.6 & 26.8 & 53.3 & 13.3 & 65.8 & 58.1 & 52.6 & 49.4 & 22.7 & 31.3 & 45.7 & 1.9 & 38.0 & 42.9 & 8.7 \\
\bottomrule
\end{tabular}%
}
\end{table*}

\subsection{Full table for Long-Context Video Modeling Evaluation}
The full results for long-context video modeling evaluation are provided in Table~\ref{tab:modality_comparison_full}.

\begin{table*}[tb]
\centering
\caption{Full Table for Comparison of model performance across Multi-Image and Video modalities.}
\label{tab:modality_comparison_full}
\resizebox{\textwidth}{!}{%
\begin{tabular}{l c cccccccccc c}
\toprule
\textbf{Model} & \textbf{Modality} & \textbf{HC} & \textbf{OC} & \textbf{FAP} & \textbf{HOI} & \textbf{HE} & \textbf{OE} & \textbf{RD-H} & \textbf{RD-O} & \textbf{VDO} & \textbf{VSD} & \textbf{Avg. Acc (\%)} \\
\midrule
Gemini-3.1-flash-lite & Multi-Image & 59.0 & 31.6 & 68.9 & 26.8 & 58.5 & 70.2 & 82.7 & 70.8 & 32.3 & 27.8 & 62.6 \\
Gemini-3.1-flash-lite & Video       & 51.8 & 30.3 & 68.9 & 17.1 & 52.8 & 55.3 & 78.6 & 66.9 & 25.8 & 27.8 & 57.6 \\
GPT-5-mini-2025-08-07 & Multi-Image & 61.4 & 42.1 & 65.1 & 34.1 & 47.2 & 67.0 & 85.7 & 77.3 & 25.8 & 36.1 & 64.6 \\
GPT-5-mini-2025-08-07 & Video       & 67.5 & 36.8 & 65.1 & 24.4 & 47.2 & 66.0 & 84.2 & 76.0 & 29.0 & 36.1 & 63.7 \\
\bottomrule
\end{tabular}%
}
\end{table*}

\subsection{Full Table for Direct Measurement of Visual Self-Awareness}
The full results for direct measurement of visual self-awareness are provided in Table~\ref{tab:exploration_metrics_full}.

\begin{table*}[tb]
\centering
\caption{Direct Measurement of Visual Self-Awareness Full Statistics.}
\label{tab:exploration_metrics_full}
\setlength{\tabcolsep}{4pt}
\resizebox{\textwidth}{!}{
\begin{tabular}{l cccccccccc cccc}
\toprule
\textbf{Model} & \textbf{HC} & \textbf{OC} & \textbf{FAP} & \textbf{HOI} & \textbf{HE} & \textbf{OE} & \textbf{RD-H} & \textbf{RD-O} & \textbf{VDO} & \textbf{VSD} & \textbf{Avg. Acc (\%)} & \textbf{Avg.} & \textbf{Avg.} & \textbf{Active} \\
& & & & & & & & & & & \textbf{(Final Step)} & \textbf{Steps} & \textbf{Reward} & \textbf{Rate (\%)} \\
\midrule
Gemini-3.1-flash-lite & 28.9 & 19.7 & 45.6 & 16.7 & 71.2 & 40.3 & 42.6 & 35.6 & 21.3 & 38.5 & 37.3 & 1.4 & 33.0 & 16.2 \\
GPT-5-mini-2025-08-07 & 32.4 & 20.5 & 50.8 & 10.0 & 59.5 & 49.2 & 51.6 & 46.1 & 32.0 & 34.4 & 41.4 & 1.4 & 36.4 & 26.3 \\
\bottomrule
\end{tabular}
}
\end{table*}

\subsection{Current View Only Evaluation}
We include a Current-View-Only evaluation, in which previously observed images are removed at each turn while the question, answer options, and textual interaction history remain available. The results in Table~\ref{tab:history_ablation} reveal that removing the visual history degrades overall accuracy.

\begin{table*}[tb]
\centering
\caption{Impact of interaction history visibility on active exploration performance.}
\label{tab:history_ablation}
\setlength{\tabcolsep}{4pt}
\resizebox{\textwidth}{!}{
\begin{tabular}{l c cccccccccc cccc}
\toprule
\textbf{Model} & \textbf{History} & \textbf{HC} & \textbf{OC} & \textbf{FAP} & \textbf{HOI} & \textbf{HE} & \textbf{OE} & \textbf{RD-H} & \textbf{RD-O} & \textbf{VDO} & \textbf{VSD} & \textbf{Avg.} & \textbf{Avg.} & \textbf{Avg.} & \textbf{Active} \\
& & & & & & & & & & & & \textbf{Acc (\%)} & \textbf{Steps} & \textbf{Reward} & \textbf{Rate (\%)} \\
\midrule
Gemini-3.1-flash-lite & Yes & 29.8 & 21.3 & 47.2 & 14.4 & 68.5 & 40.3 & 44.6 & 38.9 & 18.7 & 37.5 & 38.0 & 1.5 & 31.5 & 24.8 \\
Gemini-3.1-flash-lite & No  & 21.3 & 7.9  & 48.2 & 12.2 & 69.4 & 40.3 & 47.1 & 38.9 & 20.0 & 34.4 & 36.2 & 2.0 & 32.1 & 23.3 \\
GPT-5-mini-2025-08-07 & Yes & 34.2 & 23.6 & 50.3 & 10.0 & 59.5 & 50.8 & 49.7 & 46.7 & 28.0 & 34.4 & 41.4 & 1.4 & 34.0 & 29.4 \\
GPT-5-mini-2025-08-07 & No  & 26.7 & 29.9 & 53.8 & 12.2 & 64.0 & 47.6 & 49.7 & 43.9 & 26.7 & 33.3 & 41.0 & 1.7 & 35.7 & 25.7 \\
\bottomrule
\end{tabular}
}
\end{table*}

\subsection{Confidence Interval Analysis}
To rigorously assess statistical significance, we calculate the 95\% confidence intervals (using the Wald method) for overall accuracy across both the Active and Passive settings, including the evaluated open-source models, as shown in Table~\ref{tab:confidence_intervals}. The analysis demonstrates that the 0.3\% accuracy shift observed in the expanded view selection evaluation falls well within the margin of error. This confirms our assertion that performance remains virtually unchanged, allowing us to attribute this minor variance to statistical noise and maintain focus on the resulting efficiency reward gains.

\begin{table*}[tb]
\centering
\caption{95\% Confidence Intervals (Wald method) for overall accuracy across Active and Passive settings.}
\label{tab:confidence_intervals}
\resizebox{\textwidth}{!}{
\begin{tabular}{l ccc ccc}
\toprule
& \multicolumn{3}{c}{\textbf{Active Setting}} & \multicolumn{3}{c}{\textbf{Passive Setting}} \\
\cmidrule(lr){2-4} \cmidrule(lr){5-7}
\textbf{Model} & \textbf{Accuracy (\%)} & \textbf{Margin (\%)} & \textbf{95\% CI} & \textbf{Accuracy (\%)} & \textbf{Margin (\%)} & \textbf{95\% CI} \\
\midrule
Gemini-3.5-flash & 56.76 & $\pm$2.48 & [54.28, 59.24] & 69.20 & $\pm$3.07 & [66.13, 72.26] \\
Gemini-3.1-flash-lite & 38.05 & $\pm$2.43 & [35.62, 40.47] & 62.64 & $\pm$3.21 & [59.43, 65.86] \\
Gemini-2.5-flash & 45.41 & $\pm$2.49 & [42.92, 47.90] & 58.51 & $\pm$3.27 & [55.23, 61.78] \\
Gemini-2.5-flash-lite & 28.08 & $\pm$2.25 & [25.83, 30.33] & 45.06 & $\pm$3.31 & [41.75, 48.36] \\
GPT-5.2-2025-12-11 & 46.45 & $\pm$2.50 & [43.95, 48.94] & 62.99 & $\pm$3.21 & [59.78, 66.20] \\
GPT-5-mini-2025-08-07 & 41.43 & $\pm$2.46 & [38.97, 43.90] & 64.60 & $\pm$3.18 & [61.42, 67.78] \\
GPT-5-nano-2025-08-07 & 37.98 & $\pm$2.43 & [35.55, 40.41] & 55.40 & $\pm$3.30 & [52.10, 58.71] \\
LLaVA-OneVision-2-8B & 36.09 & $\pm$2.40 & [33.69, 38.49] & 58.85 & $\pm$3.27 & [55.58, 62.12] \\
Qwen3.5-9B-No-Thinking & 34.53 & $\pm$2.38 & [32.15, 36.91] & - & - & - \\
\bottomrule
\end{tabular}
}
\end{table*}

\subsection{Extended Open-Source Model Evaluation}
\label{app:more_open_source}
Comprehensive results for additional open-source models are detailed in Table~\ref{tab:opensource_results}. Inference latency is benchmarked using two NVIDIA RTX 3090 GPUs for standard models and a single NVIDIA H100 GPU for larger models, such as 27B and 35B parameters. While baseline inference typically concludes within a few hours, models employing explicit reasoning (``thinking'') mechanisms suffer severe latency overheads during multi-turn Active exploration (more than 20 hours). Furthermore, we attempted to evaluate larger open-source variants in the Active setting; however, the iterative view-exploration process incurred computationally prohibitive inference times (exceeding several days), precluding their inclusion in the final benchmark.
\begin{table*}[t]
\centering
\caption{Comprehensive Evaluation Results of Open-Source Vision-Language Models on SATE-Bench across Active (Multi-turn Spatial Exploration) and Passive (Single-turn Static Observation) settings. Task abbreviations: \textbf{HC} (Human Counting), \textbf{OC} (Object Counting), \textbf{FAP} (Fine-grained Action \& Pose), \textbf{HOI} (Human-Object Interaction), \textbf{HE} (Human Existence), \textbf{OE} (Object Existence), \textbf{RD-H} (Relative Distance to Human), \textbf{RD-O} (Relative Distance to Object), \textbf{VDO} (View-Specific Depth Ordering), \textbf{VSD} (View-Specific Direction). In the Passive setting, multi-turn exploration metrics are marked with ``$-$''. Best results among open-source models are bolded.}
\label{tab:opensource_results}
\resizebox{\textwidth}{!}{%
\begin{tabular}{l|cccccccccc|cccc}
\toprule
\textbf{Model} & \textbf{HC} & \textbf{OC} & \textbf{FAP} & \textbf{HOI} & \textbf{HE} & \textbf{OE} & \textbf{RD-H} & \textbf{RD-O} & \textbf{VDO} & \textbf{VSD} & \textbf{Avg.} & \textbf{Avg.} & \textbf{Avg.} & \textbf{Active} \\
& & & & & & & & & & & \textbf{Acc (\%)} & \textbf{Steps} & \textbf{Reward} & \textbf{Rate (\%)} \\
\midrule
\multicolumn{15}{l}{\textit{Active Setting (Multi-turn Spatial Exploration)}} \\
\midrule
Qwen2.5-VL-3B-Instruct       & 17.8 & 7.5 & 34.9 & 14.3 & 61.3 & 41.9 & 36.2 & 36.7 & 23.0 & 21.9 & 30.4 & 1.0 & 27.5 & 0.2 \\
Qwen2.5-VL-7B-Instruct       & 20.0 & 11.8 & 23.6 & 11.1 & 66.7 & 37.9 & 34.6 & 36.7 & 18.7 & 33.3 & 29.8 & 1.0 & 26.9 & 0.4 \\
Qwen3-VL-8B-Instruct         & 9.3 & 9.4 & 27.2 & 11.1 & 58.6 & 37.1 & 44.6 & \textbf{44.4} & 20.0 & 31.2 & 30.7 & 1.6 & 27.1 & \textbf{15.6} \\
Qwen3-VL-8B-Thinking         & 12.0 & 11.8 & 19.5 & 8.9 & \textbf{72.1} & 44.4 & 42.3 & 30.6 & 18.7 & 34.4 & 29.8 & 1.1 & 26.7 & 5.8 \\
Qwen3.5-9B                   & \textbf{27.1} & 12.2 & \textbf{40.0} & 18.8 & 65.0 & 47.6 & 47.1 & 38.9 & 25.3 & 29.2 & \textbf{36.7} & 1.1 & \textbf{33.1} & 4.1 \\
Qwen3.5-9B-No-Thinking          & 24.0 & 12.6 & 38.5 & 12.2 & 62.2 & 44.4 & 41.3 & 39.4 & 16.0 & 39.6 & 34.5 & 1.0 & 31.2 & 0.2 \\
Gemma-3-4B-it                & 25.3 & \textbf{17.7} & 31.8 & \textbf{26.8} & 32.8 & 52.4 & 32.7 & 40.0 & 11.5 & 33.3 & 31.0 & 1.1 & 28.0 & 3.3 \\
LLaVA-OneVision-2-8B           & 24.0 & 10.2 & 32.3 & 10.0 & 64.9 & 41.9 & \textbf{47.8} & 43.9 & \textbf{29.3} & \textbf{42.7} & 36.1 & 1.0 & 32.7 & 0.0 \\
\midrule
\multicolumn{15}{l}{\textit{Passive Setting (Single-turn Static Observation)}} \\
\midrule
Qwen2.5-VL-3B-Instruct       & 49.4 & 28.2 & 51.9 & 22.6 & 62.1 & 55.3 & 65.3 & 58.4 & 21.6 & 33.3 & 50.9 & - & - & - \\
Qwen2.5-VL-7B-Instruct       & 49.4 & 31.6 & 56.6 & 22.0 & 62.3 & 55.3 & 67.9 & 70.8 & 19.4 & 30.6 & 54.9 & - & - & - \\
Qwen2.5-VL-32B-Instruct      & 50.6 & 36.8 & 61.3 & 24.4 & 64.2 & 59.6 & 79.1 & 71.4 & 25.8 & 36.1 & 59.9 & - & - & - \\
Qwen3-VL-2B-Instruct         & 41.0 & 30.3 & 55.7 & 22.0 & 50.9 & 53.2 & 65.8 & 62.3 & 32.3 & 30.6 & 51.5 & - & - & - \\
Qwen3-VL-4B-Instruct         & 45.8 & 21.1 & 57.5 & 24.4 & 39.6 & 54.3 & 76.0 & 70.1 & \textbf{38.7} & 30.6 & 54.8 & - & - & - \\
Qwen3-VL-4B-Thinking         & 45.8 & 23.7 & 55.7 & 19.5 & \textbf{69.8} & 66.0 & 71.9 & 63.6 & 22.6 & 25.0 & 54.8 & - & - & - \\
Qwen3-VL-8B-Instruct         & 53.0 & 27.6 & 57.5 & 19.5 & 52.8 & 54.3 & 74.0 & 68.2 & 25.8 & 36.1 & 55.6 & - & - & - \\
Qwen3-VL-8B-Thinking         & 55.4 & 30.3 & 61.3 & 24.4 & 62.3 & 62.8 & 73.5 & 67.5 & 25.8 & 27.8 & 57.7 & - & - & - \\
Qwen3.5-9B                   & 53.0 & 35.3 & 62.3 & 22.6 & 48.5 & 58.5 & 77.0 & 66.2 & 32.4 & \textbf{38.9} & 56.9 & - & - & - \\
Qwen3.5-27B                  & 45.8 & 38.2 & 69.8 & 34.1 & 52.8 & 62.8 & \textbf{81.1} & 74.7 & 32.3 & 22.2 & 61.4 & - & - & - \\
Qwen3.5-35B-A3B              & 51.8 & \textbf{42.1} & \textbf{71.7} & \textbf{41.5} & 52.8 & 67.0 & 78.6 & 76.0 & 35.5 & 33.3 & \textbf{63.6} & - & - & - \\
Qwen3.6-27B                  & 51.8 & 28.9 & 69.8 & 39.0 & 60.4 & \textbf{68.1} & 80.6 & \textbf{76.6} & 35.5 & 27.8 & 63.0 & - & - & - \\
Qwen3.6-35B-A3B              & \textbf{60.2} & 39.5 & 70.8 & 36.6 & 50.9 & 62.8 & 79.6 & 75.3 & 32.3 & 25.0 & 62.9 & - & - & - \\
Gemma-3-4B-it                & 33.7 & 25.9 & 37.7 & 26.4 & 25.8 & 54.3 & 49.0 & 51.9 & 24.3 & 25.0 & 40.2 & - & - & - \\
LLaVA-OneVision-2-8B           & 56.6 & 31.6 & 62.3 & 26.8 & 52.8 & 54.3 & 80.1 & 71.4 & 25.8 & 27.8 & 58.9 & - & - & - \\
\bottomrule
\end{tabular}%
}
\end{table*}

\subsection{Exploration Details}
SATE-Bench evaluates whether MLLMs can actively acquire missing visual evidence in severely occluded human-centric scenes. We utilize closed-loop camera layouts that provide complete azimuthal coverage, framing the task as controlled exocentric view selection rather than unconstrained 3D navigation. The active exploration environment details can be found below: (1) the maximum number of exploration steps is strictly capped at the total available views per scene (typically eight); (2) models may freely revisit perspectives without artificial restrictions on trajectory loops; and (3) by default, models maintain a complete visual history of all previously queried images to facilitate cross-view spatial fusion.

\section{Abstention Option Analysis}
For the 5-option QA pairs, the Abstention Option is the ground-truth answer in 2.48\% of Active samples and 1.84\% of Passive samples. As shown in Table~\ref{tab:abstention_rates}, human evaluators select it in fewer than 2\% of cases, closely matching its ground-truth frequency. In contrast, weaker MLLMs frequently over-select the Abstention Option under severe occlusion, whereas stronger models tend to under-select it and instead guess a concrete answer when visual evidence remains insufficient. These results reveal both excessive abstention reliance and overconfident hallucination. In the fully observable Passive setting, Abstention Option selection decreases across all models as visual ambiguity is reduced.

\begin{table*}[tb]
\centering
\caption{Selection rate and accuracy for the Abstention Option across Active and Passive settings.}
\label{tab:abstention_rates}
\resizebox{\textwidth}{!}{
\begin{tabular}{lcccc}
\toprule
\textbf{Model} & \textbf{Abstention Selection Rate (Active)} & \textbf{Accuracy on Abstention (Active)} & \textbf{Abstention Selection Rate (Passive)} & \textbf{Accuracy on Abstention (Passive)} \\
\midrule
Gemini-2.5-flash & 14.57\% & 47.37\% & 2.24\% & 18.75\% \\
Gemini-2.5-flash-lite & 54.38\% & 92.11\% & 16.36\% & 75.00\% \\
Gemini-3.1-flash-lite & 22.25\% & 71.05\% & 2.10\% & 18.75\% \\
Gemini-3.5-flash & 2.50\% & 0.00\% & 0.98\% & 12.50\% \\
GPT-5-mini-2025-08-07 & 5.62\% & 18.42\% & 0.56\% & 6.25\% \\
GPT-5-nano-2025-08-07 & 12.41\% & 36.84\% & 0.70\% & 18.75\% \\
GPT-5.2-2025-12-11 & 11.55\% & 31.58\% & 0.98\% & 6.25\% \\
LLaVA-OneVision-2-8B & 25.14\% & 60.53\% & 2.94\% & 31.25\% \\
Qwen3.5-9B & 20.53\% & 15.79\% & 1.40\% & 12.50\% \\
\bottomrule
\end{tabular}
}
\end{table*}

\section{Ablation Study of Average Reward Coefficient}
We conduct a sensitivity analysis using $\lambda \in \{0.05, 0.1, 0.2, 0.5\}$. As shown in Table~\ref{tab:sensitivity_lambda}, increasing $\lambda$ reduces the absolute reward values, while the main qualitative conclusions remain stable under moderate penalties from 0.05 to 0.2. A large value of $\lambda=0.5$ overemphasizes short trajectories and changes the model rankings. We therefore retain $\lambda=0.1$, which applies a moderate penalty of approximately 9.5\% per additional exploration step, as a balanced trade-off between accuracy and exploration efficiency.

\begin{table}[tb]
\centering
\caption{Sensitivity analysis of the step penalty parameter $\lambda$ on average reward. The reported Average Reward is scaled by 100.}
\label{tab:sensitivity_lambda}
\setlength{\tabcolsep}{3pt}
\resizebox{\columnwidth}{!}{
\begin{tabular}{lcccc}
\toprule
Model & $\lambda=0.05$ & $\lambda=0.1$ & $\lambda=0.2$ & $\lambda=0.5$ \\
\midrule
Gemini-3.5-flash      & 45.85 & 40.60 & 32.25 & 17.48 \\
Gemini-3.1-flash-lite & 33.67 & 31.45 & 27.56 & 19.04 \\
Gemini-2.5-flash      & 38.69 & 35.31 & 29.75 & 18.97 \\
Gemini-2.5-flash-lite & 26.15 & 24.49 & 21.51 & 14.83 \\
GPT-5.2-2025-12-11    & 39.56 & 36.50 & 31.15 & 19.80 \\
GPT-5-mini-2025-08-07 & 36.24 & 33.95 & 29.88 & 20.69 \\
GPT-5-nano-2025-08-07 & 33.93 & 32.01 & 28.54 & 20.40 \\
\bottomrule
\end{tabular}
}
\end{table}

\section{Training Details}
\label{app:train_details}

\subsection{Preliminaries}
We first revisit Group Relative Policy Optimization (GRPO), an efficient variant of Proximal Policy Optimization (PPO) that eliminates the critic model. Given a multimodal dataset $\mathcal{D}$ containing questions $q$, ground-truth answers $y$, and visual inputs $\mathcal{I}$, GRPO encourages accurate responses by evaluating a relative group of candidate outputs. 

For a policy $\pi_{\theta}$, we sample a group of $G$ sequences $\{o_i\}_{i=1}^G$ from the old policy $\pi_{\theta_{\rm old}}(\cdot \mid q, \mathcal{I})$. To accommodate single-column width constraints, we decompose the GRPO objective into the core expectation and the token-level loss $\mathcal{L}_{i,t}(\theta)$:

\begin{align}
    \mathcal{J}_{\rm GRPO}(\theta) &= \mathbb{E} \left[ \frac{1}{G} \sum_{i=1}^{G} \frac{1}{|o_i|} \sum_{t=1}^{|o_i|} \mathcal{L}_{i,t}(\theta) \right], \\
    \mathcal{L}_{i,t}(\theta) &= \min \Bigl( r_{i,t}(\theta)\hat{A}_{i, t}, \text{clip}\bigl(r_{i,t}(\theta), 1-\epsilon, 1+\epsilon\bigr) \hat{A}_{i, t} \Bigr) \nonumber \\
    &\quad - \lambda D_{\rm KL}\left(\pi_\theta \parallel \pi_{\rm ref}\right),
\end{align}

where the token probability ratio $r_{i,t}(\theta)$ is defined as:
\begin{equation}
    r_{i,t}(\theta) = \frac{\pi_{\theta}(o_{i,t} \mid q, \mathcal{I}, o_{i, <t})}{\pi_{\theta_{\rm old}}(o_{i,t} \mid q, \mathcal{I}, o_{i, <t})}.
\end{equation}

Here, $\epsilon$ restricts the policy update bounds, and $\lambda$ scales the KL penalty against a reference policy $\pi_{\rm ref}$ to ensure optimization stability. The token-level advantage $\hat{A}_{i,t}$ is computed by normalizing the sequence-level reward $R_i$ across the sampled sequence group $\mathbf{R} = \{R_1, \dots, R_G\}$:

\begin{equation}
    \hat{A}_{i,t} = \frac{R_i - \text{mean}(\mathbf{R})}{\text{std}(\mathbf{R})}.
\end{equation}

The design of the reward $R_i$ is highly flexible; typically, rule-based verifiable rewards are employed to evaluate accuracy and format compliance against the ground truth $y$.

\subsection{Training Data Details}
We construct our training corpus by sampling 86 scenes from the Ego-Exo4D~\cite{grauman2024ego} and M3GYM~\cite{xu2025m3gym} datasets, yielding a total of 19,370 question-answer pairs. This dataset is randomly partitioned into a 90\% training set and a 10\% validation set. Detailed text length statistics are provided in Table~\ref{tab:train_val_text_stats}.

\begin{table}[h]
\centering
\caption{Text Length Statistics for Combined Training and Validation Sets ($N=19,370$).}
\label{tab:train_val_text_stats}
\begin{tabular}{lc}
\toprule
\textbf{Metric} & \textbf{Value} \\
\midrule
Average Question Length & 119.63 \\
Max Question Length & 344.00 \\
Average Option Length & 37.48 \\
Max Option Length & 185.00 \\
Average `Purely Text' Option Length & 45.03 \\
Average Correct Answer Length & 33.74 \\
Max Correct Answer Length & 185.00 \\
\bottomrule
\end{tabular}
\end{table}

\subsection{Training Hyperparameters}
Table~\ref{tab:grpo_hyperparameters} shows the key training hyperparameters.

\begin{table}[tb]
\centering
\caption{Key hyperparameters for GRPO fine-tuning.}
\label{tab:grpo_hyperparameters}
\resizebox{\columnwidth}{!}{
\begin{tabular}{lc}
\toprule
\textbf{Hyperparameter} & \textbf{Value} \\
\midrule
\multicolumn{2}{c}{\textit{Training \& Optimization}} \\
\midrule
Base Model & Qwen2.5-VL-3B-Instruct \\
Learning Rate & $1 \times 10^{-5}$ \\
Train Batch Size (per device) & 2 \\
Gradient Accumulation Steps & 16 \\
Max Training Steps & 300 \\
Warmup Steps & 30 \\
\midrule
\multicolumn{2}{c}{\textit{GRPO Specifics}} \\
\midrule
Number of Generations ($G$) & 8 \\
Max Completion Length & 2048 \\
Temperature & 1.0 \\
KL Penalty ($\beta$) & 0.005 \\
\midrule
\multicolumn{2}{c}{\textit{Reward Formulation Weights}} \\
\midrule
Accuracy Weight & 1.0 \\
Format Weight & 0.5 \\
Action Weight & 0.2 \\
\midrule
\multicolumn{2}{c}{\textit{LoRA Configuration}} \\
\midrule
Rank ($r$) & 16 \\
Alpha ($\alpha$) & 16 \\
Dropout & 0.1 \\
Target Modules & \texttt{q, k, v, o, gate, up, down\_proj} \\
\bottomrule
\end{tabular}
}
\end{table}

\subsection{Full Table for Training}
The evaluation results of GRPO Fine-Tuned Models on SATE-Bench across Active and Passive settings can be found in Table~\ref{tab:grpo_results_full}.
\begin{table*}[t]
\centering
\caption{Evaluation Results of GRPO Fine-Tuned Models on SATE-Bench across Active (Multi-turn Spatial Exploration) and Passive (Single-turn Static Observation) settings. Task abbreviations: \textbf{HC} (Human Counting), \textbf{OC} (Object Counting), \textbf{FAP} (Fine-grained Action \& Pose), \textbf{HOI} (Human-Object Interaction), \textbf{HE} (Human Existence), \textbf{OE} (Object Existence), \textbf{RD-H} (Relative Distance to Human), \textbf{RD-O} (Relative Distance to Object), \textbf{VDO} (View-Specific Depth Ordering), \textbf{VSD} (View-Specific Direction). In the Passive setting, multi-turn metrics are marked with ``$-$''. Best results among tuned models are bolded.}
\label{tab:grpo_results_full}
\resizebox{\textwidth}{!}{%
\begin{tabular}{l|cccccccccc|cccc}
\toprule
\textbf{Model} & \textbf{HC} & \textbf{OC} & \textbf{FAP} & \textbf{HOI} & \textbf{HE} & \textbf{OE} & \textbf{RD-H} & \textbf{RD-O} & \textbf{VDO} & \textbf{VSD} & \textbf{Avg.} & \textbf{Avg.} & \textbf{Avg.} & \textbf{Active} \\
& & & & & & & & & & & \textbf{Acc (\%)} & \textbf{Steps} & \textbf{Reward} & \textbf{Rate (\%)} \\
\midrule
\multicolumn{15}{l}{\textit{Active Setting: Base Model vs. GRPO Tuned Models}} \\
\midrule
Qwen2.5-VL-3B (Base)         & 17.8 & 7.5 & \textbf{34.9} & \textbf{14.3} & 61.3 & 41.9 & \textbf{36.2} & \textbf{36.7} & \textbf{23.0} & 21.9 & \textbf{30.4} & 1.0 & \textbf{27.5} & 0.2 \\
Qwen2.5-VL-3B-Instruct-005   & 20.9 & 8.2 & 12.8 & 5.4 & \textbf{73.0} & 44.4 & 17.3 & 14.4 & 11.5 & \textbf{30.2} & 22.5 & 4.3 & 15.0 & 49.5 \\
Qwen2.5-VL-3B-Instruct-02    & 18.7 & 7.5 & 11.8 & 7.1 & \textbf{73.0} & 46.0 & 17.6 & 15.6 & 13.8 & 24.0 & 22.2 & 4.7 & 14.6 & 55.1 \\
Qwen2.5-VL-3B-Instruct-04    & \textbf{24.9} & 7.5 & 11.8 & 7.1 & 69.3 & 40.3 & 14.7 & 13.3 & 12.6 & 28.1 & 21.7 & 4.7 & 14.0 & 56.3 \\
Qwen2.5-VL-3B-Instruct-06    & 16.4 & 6.1 & 10.3 & 3.6 & 69.3 & 43.5 & 14.1 & 15.0 & 16.1 & 24.0 & 20.2 & 5.1 & 12.6 & 60.4 \\
Qwen2.5-VL-3B-Instruct-08    & 22.7 & \textbf{8.8} & 9.2 & 4.5 & 70.8 & 48.4 & 14.1 & 13.3 & 12.6 & 25.0 & 21.5 & 5.3 & 12.8 & 65.0 \\
Qwen2.5-VL-3B-Instruct-10    & 19.6 & 8.2 & 7.7 & 2.7 & 70.8 & 41.9 & 9.3 & 6.1 & 14.9 & 21.9 & 18.4 & 5.3 & 10.4 & 63.0 \\
Qwen2.5-VL-3B-Instruct-12    & 19.6 & 5.4 & 8.2 & 1.8 & 67.9 & 47.6 & 9.6 & 5.6 & 9.2 & 19.8 & 17.9 & 6.0 & 9.7 & \textbf{72.8} \\
Qwen2.5-VL-3B-Instruct-14    & 21.3 & 6.8 & 5.6 & 2.7 & 67.9 & \textbf{49.2} & 9.9 & 5.6 & 14.9 & 22.9 & 18.7 & 5.9 & 10.0 & 72.0 \\
\midrule
\multicolumn{15}{l}{\textit{Passive Setting: Base Model vs. GRPO Tuned Models}} \\
\midrule
Qwen2.5-VL-3B (Base)         & 49.4 & 28.2 & 51.9 & 22.6 & 62.1 & 55.3 & \textbf{65.3} & 58.4 & 21.6 & 33.3 & 50.9 & - & - & - \\
Qwen2.5-VL-3B-Instruct-005   & 45.8 & 29.4 & 50.9 & 24.5 & 59.1 & \textbf{56.4} & \textbf{65.3} & \textbf{59.1} & 21.6 & \textbf{41.7} & 51.0 & - & - & - \\
Qwen2.5-VL-3B-Instruct-02    & \textbf{50.6} & 30.6 & \textbf{52.8} & 24.5 & 59.1 & \textbf{56.4} & 63.8 & 58.4 & 21.6 & 38.9 & 51.2 & - & - & - \\
Qwen2.5-VL-3B-Instruct-04    & 48.2 & 28.2 & 51.9 & 24.5 & 57.6 & \textbf{56.4} & 64.3 & \textbf{59.1} & 21.6 & \textbf{41.7} & 50.9 & - & - & - \\
Qwen2.5-VL-3B-Instruct-06    & 49.4 & 30.6 & \textbf{52.8} & \textbf{26.4} & 57.6 & \textbf{56.4} & 63.8 & 58.4 & 21.6 & \textbf{41.7} & 51.2 & - & - & - \\
Qwen2.5-VL-3B-Instruct-08    & 48.2 & 30.6 & \textbf{52.8} & 22.6 & 60.6 & 55.3 & 64.3 & \textbf{59.1} & \textbf{27.0} & 38.9 & 51.3 & - & - & - \\
Qwen2.5-VL-3B-Instruct-10    & 49.4 & \textbf{31.8} & \textbf{52.8} & 22.6 & \textbf{63.6} & 55.3 & 63.3 & \textbf{59.1} & \textbf{27.0} & \textbf{41.7} & \textbf{51.6} & - & - & - \\
Qwen2.5-VL-3B-Instruct-12    & \textbf{50.6} & 28.2 & 51.9 & 24.5 & \textbf{63.6} & 55.3 & 63.3 & \textbf{59.1} & \textbf{27.0} & \textbf{41.7} & 51.4 & - & - & - \\
Qwen2.5-VL-3B-Instruct-14    & 48.2 & 29.4 & \textbf{52.8} & 22.6 & \textbf{63.6} & \textbf{56.4} & 63.8 & \textbf{59.1} & \textbf{27.0} & 38.9 & 51.4 & - & - & - \\
\bottomrule
\end{tabular}%
}
\end{table*}

\section{Computational Resources Analysis}
Evaluation of the smaller open-source models is conducted on two NVIDIA RTX 3090 GPUs, while larger models utilize a single NVIDIA H100 GPU. All GRPO fine-tuning procedures are executed on a single compute node equipped with four NVIDIA H100 GPUs, with training runs typically concluding within a few hours.

For closed-source models, inference is performed via the official Gemini and OpenAI GPT APIs. The iterative nature of the Active setting results in higher API costs, exceeding \$150 USD for the largest proprietary models, whereas smaller models and the standard Passive setting generally incur costs under \$50 USD. 

\section{Prompt Details}
\label{sec:prompts}
In this section, we provide the complete prompt templates utilized throughout our evaluations, including both the system prompts and the user instructions. The exact formatting and instructions are detailed in Figure~\ref{fig:prompt_passive} for the Passive setting, Figure~\ref{fig:prompt_active} for the Active setting, Figure~\ref{fig:prompt_active_ablation} for the expanded view selection evaluation, Figure~\ref{fig:prompt_active_confidence} for confidence score evaluation, and Figure~\ref{fig:prompt_active_exhaustive} for active exhaustive search evaluation.

\section{Limitations}
First, our spatial querying interface is currently constrained to a discrete ring (left/right camera shifts). Second, we rely on synchronized static frames, leaving active temporal exploration, such as dynamic video rewinding, unexamined. Finally, iterative multi-turn querying incurs a high computational cost compared to single-pass evaluations. This latency underscores an urgent need for memory-efficient multi-view caching mechanisms to sustainably process expanding visual contexts during continuous perspective exploration.

\begin{figure*}[t]
\begin{tcolorbox}[
title=Prompts for Passive Evaluation,
colback=white,
colframe=black!70,
fonttitle=\scriptsize\bfseries
]
{\footnotesize\textbf{System Prompt.}}\vspace{-0.35em}

\begin{tcolorbox}[
colback=black!2,
colframe=black!10,
boxrule=0.25pt,
arc=0.5mm,
left=0.45mm,
right=0.45mm,
top=0.1mm,
bottom=0.1mm
]
{\scriptsize\ttfamily
\setlength{\baselineskip}{8.2pt}
You are an expert in multi-view humans and objects understanding. You will be provided with several views of a scene. Carefully analyze the images based on the user's question, then provide the final answer as a single letter corresponding to the correct option. You must answer strictly with one single letter from [A, B, C, D, E]. Do not add any explanation or extra text. Directly answer with a single letter!
}
\end{tcolorbox}

{\footnotesize\textbf{User Instruction Input Prompt for 5-Option Question.}}\vspace{-0.35em}

\begin{tcolorbox}[
colback=black!2,
colframe=black!10,
boxrule=0.25pt,
arc=0.5mm,
left=0.45mm,
right=0.45mm,
top=0.1mm,
bottom=0.1mm
]
{\scriptsize\ttfamily
\setlength{\baselineskip}{7.5pt}
You are given multiple synchronized views of the same scene.\\

Question: \{question\}\\

Options: A) \{optA\} \; B) \{optB\} \; C) \{optC\} \; D) \{optD\} \; E) \{optE\}\\

Output:
Answer with exactly ONE letter from [A, B, C, D, E], e.g., just "C". No explanation. Only one letter.
}
\end{tcolorbox}

{\footnotesize\textbf{User Instruction Input Prompt for 2-Option Question.}}\vspace{-0.35em}

\begin{tcolorbox}[
colback=black!2,
colframe=black!10,
boxrule=0.25pt,
arc=0.5mm,
left=0.45mm,
right=0.45mm,
top=0.1mm,
bottom=0.1mm
]
{\scriptsize\ttfamily
\setlength{\baselineskip}{7.5pt}
You are given multiple synchronized views of the same scene.\\

Question: \{question\}\\

Options: A) \{optA\} \; B) \{optB\}\\

Output:
Answer with exactly ONE letter from [A, B], e.g., just "B". No explanation. Only one letter.
}
\end{tcolorbox}
\end{tcolorbox}
\caption{Prompts for Passive Evaluation}
\label{fig:prompt_passive}
\end{figure*}

\begin{figure*}[t]
\begin{tcolorbox}[
title=Prompts for Active Evaluation,
colback=white,
colframe=black!70,
fonttitle=\scriptsize\bfseries
]
{\footnotesize\textbf{System Prompt.}}\vspace{-0.35em}

\begin{tcolorbox}[
colback=black!2,
colframe=black!10,
boxrule=0.25pt,
arc=0.5mm,
left=0.45mm,
right=0.45mm,
top=0.1mm,
bottom=0.1mm
]
{\scriptsize\ttfamily
\setlength{\baselineskip}{8.2pt}
You are an expert in multi-view humans and objects understanding. You are currently viewing 1 out of \{num\_views\} possible camera viewpoints of a single scene. These viewpoints are uniformly distributed in a circle around the scene and sorted in clockwise order.\\

Your task is to answer the user's multiple-choice question by carefully analyzing the provided image. You must make a strict decision between answering the question or exploring the scene further:\\

1. ANSWER: If the current view provides sufficient information to confidently answer the question, output the correct option.\\
2. EXPLORE: If the current view lacks crucial information, is heavily occluded, or
if the question requires full-scene context that this single angle cannot provide, you must request a new viewpoint.\\
   - Option 'X': Move to the adjacent left (clockwise) viewpoint.\\
   - Option 'Y': Move to the adjacent right (counter-clockwise) viewpoint.\\

CRITICAL CONSTRAINT:
Your final output must be exactly ONE single letter from the set [A, B, C, D, E, X, Y]. 
Do not include any explanations, formatting, punctuation, or extra text. Directly output the single letter.
}
\end{tcolorbox}

{\footnotesize\textbf{User Instruction Input Prompt for 5-Option Question.}}\vspace{-0.35em}

\begin{tcolorbox}[
colback=black!2,
colframe=black!10,
boxrule=0.25pt,
arc=0.5mm,
left=0.45mm,
right=0.45mm,
top=0.1mm,
bottom=0.1mm
]
{\scriptsize\ttfamily
\setlength{\baselineskip}{7.5pt}
You are given multiple synchronized views of the same scene.\\

Question: \{question\}\\

Options: A) \{optA\} \; B) \{optB\} \; C) \{optC\} \; D) \{optD\} \; E) \{optE\}\\

Exploration Options: X) \{optX\} \; Y) \{optY\} \\

Output:
Answer with exactly ONE letter from [A, B, C, D, E, X, Y], e.g., just "C". No explanation. Only one letter.
}
\end{tcolorbox}

{\footnotesize\textbf{User Instruction Input Prompt for 2-Option Question.}}\vspace{-0.35em}

\begin{tcolorbox}[
colback=black!2,
colframe=black!10,
boxrule=0.25pt,
arc=0.5mm,
left=0.45mm,
right=0.45mm,
top=0.1mm,
bottom=0.1mm
]
{\scriptsize\ttfamily
\setlength{\baselineskip}{7.5pt}
You are given multiple synchronized views of the same scene.\\

Question: \{question\}\\

Options: A) \{optA\} \; B) \{optB\}\\

Exploration Options: X) \{optX\} \; Y) \{optY\} \\

Output:
Answer with exactly ONE letter from [A, B, X, Y], e.g., just "B". No explanation. Only one letter.
}
\end{tcolorbox}
\end{tcolorbox}
\caption{Prompts for Active Evaluation}
\label{fig:prompt_active}
\end{figure*}

\begin{figure*}[t]
\begin{tcolorbox}[
title=Prompts for Active Evaluation in Ablation Study,
colback=white,
colframe=black!70,
fonttitle=\scriptsize\bfseries
]
{\footnotesize\textbf{System Prompt.}}\vspace{-0.35em}

\begin{tcolorbox}[
colback=black!2,
colframe=black!10,
boxrule=0.25pt,
arc=0.5mm,
left=0.45mm,
right=0.45mm,
top=0.1mm,
bottom=0.1mm
]
{\scriptsize\ttfamily
\setlength{\baselineskip}{8.2pt}
You are an expert in multi-view humans and objects understanding. You are currently viewing 1 out of \{num\_views\} possible camera viewpoints of a single scene. These viewpoints are uniformly distributed in a circle around the scene and sorted in clockwise order.\\

Your task is to answer the user's multiple-choice question by carefully analyzing the provided image. You must make a strict decision between answering the question or exploring the scene further:\\

1. ANSWER: If the current view provides sufficient information to confidently answer the question, output the correct option.\\
2. EXPLORE: If the current view lacks crucial information, is heavily occluded, or
if the question requires full-scene context that this single angle cannot provide, you must request a new viewpoint.\\
   - Option 'W': Move 2 viewpoints to the left (clockwise).\\
   - Option 'X': Move 1 viewpoint to the adjacent left (clockwise).\\
   - Option 'Y': Move 1 viewpoint to the adjacent right (counter-clockwise).\\
   - Option 'Z': Move 2 viewpoints to the right (counter-clockwise).\\

CRITICAL CONSTRAINT:
Your final output must be exactly ONE single letter from the set [A, B, C, D, E, W, X, Y, Z]. 
Do not include any explanations, formatting, punctuation, or extra text. Directly output the single letter.
}
\end{tcolorbox}

{\footnotesize\textbf{User Instruction Input Prompt for 5-Option Question.}}\vspace{-0.35em}

\begin{tcolorbox}[
colback=black!2,
colframe=black!10,
boxrule=0.25pt,
arc=0.5mm,
left=0.45mm,
right=0.45mm,
top=0.1mm,
bottom=0.1mm
]
{\scriptsize\ttfamily
\setlength{\baselineskip}{7.5pt}
You are given multiple synchronized views of the same scene.\\

Question: \{question\}\\

Options: A) \{optA\} \; B) \{optB\} \; C) \{optC\} \; D) \{optD\} \; E) \{optE\}\\

Exploration Options: W) \{optW\} \; X) \{optX\} \; Y) \{optY\} \; Z) \{optZ\} \\

Output:
Answer with exactly ONE letter from [A, B, C, D, E, W, X, Y, Z], e.g., just "C". No explanation. Only one letter.
}
\end{tcolorbox}

{\footnotesize\textbf{User Instruction Input Prompt for 2-Option Question.}}\vspace{-0.35em}

\begin{tcolorbox}[
colback=black!2,
colframe=black!10,
boxrule=0.25pt,
arc=0.5mm,
left=0.45mm,
right=0.45mm,
top=0.1mm,
bottom=0.1mm
]
{\scriptsize\ttfamily
\setlength{\baselineskip}{7.5pt}
You are given multiple synchronized views of the same scene.\\

Question: \{question\}\\

Options: A) \{optA\} \; B) \{optB\}\\

Exploration Options: W) \{optW\} \; X) \{optX\} \; Y) \{optY\} \; Z) \{optZ\} \\

Output:
Answer with exactly ONE letter from [A, B, W, X, Y, Z], e.g., just "B". No explanation. Only one letter.
}
\end{tcolorbox}
\end{tcolorbox}
\caption{Prompts for Expanded View Selection Evaluation}
\label{fig:prompt_active_ablation}
\end{figure*}

\section{Consistency with Intended Artifact Use}
\label{sec:artifact_use}

In developing SATE-Bench, we utilized ten existing multi-view human-centric datasets (detailed in Appendix~\ref{sec:source_dataset_details}). Our employment of these artifacts is strictly consistent with their original intended use: advancing academic research in human-centric visual perception, 3D geometry, and spatial reasoning. Our evaluation of foundation models falls entirely within this designated academic scope.

For the artifact we create and release (SATE-Bench), the intended use is strictly limited to non-commercial, academic research focused on evaluating the spatial reasoning and active view-exploration capabilities of MLLMs. We explicitly specify that this intended use is fully compatible with the original access conditions of all source datasets. Because SATE-Bench functions as a derivative work of data originally accessed solely for research purposes, we stipulate that SATE-Bench, including all its annotations, prompt templates, and derived multi-view frames, must not be used outside of research contexts or for commercial applications. Our release will enforce these constraints, strictly adhering to the most restrictive non-commercial licensing terms of the underlying source materials.

\section{Personally Identifying Information and Offensive Content}
\label{sec:pii_and_offensive_content}

In the curation of SATE-Bench, we implemented strict protocols to identify and mitigate the presence of Personally Identifying Information (PII) and offensive content. First, our benchmark is entirely derived from established, peer-reviewed public datasets (e.g., M3GYM, Ego-Exo4D, 4D-OR) that have already undergone their respective Institutional Review Board (IRB) approvals and subject consent processes prior to their initial public release.

During our multi-stage annotation and quality control pipeline, all 2,525 Question-Answer pairs and their corresponding multi-view frames are subjected to rigorous manual review by our annotation team. To protect individual identities, annotators are strictly prohibited from utilizing names, specific facial features, or any uniquely identifying metadata in the prompts, distractors, or ground-truth answers. Instead, all human subjects are referenced exclusively through generic physical descriptors, clothing attributes, or geometric spatial relationships, such as ``the person wearing a red shirt'' and ``the individual closest to the surgical table''. 

Furthermore, annotators are explicitly trained to flag and filter any frames or generated text containing offensive, harmful, or inappropriate content. Given that the source domains exclusively encompass structured environments, the inherent risk of offensive material is exceptionally low. Nevertheless, our independent cross-validation phase served as a final safeguard, ensuring that all finalized benchmark materials are completely anonymized and adhere strictly to academic safety and ethical standards.

\section{Documentation of Artifacts}
\label{sec:artifact_documentation}

In accordance with robust artifact documentation standards, we provide the following metadata regarding the composition and scope of SATE-Bench:

\noindent\textbf{Coverage of Domains:} The benchmark encompasses 316 real-world, physical environments. These domains are strictly structured and include indoor living spaces, athletic facilities, clinical operating rooms, and outdoor activities. The visual focus is exclusively on dense, multi-person spatial interactions, complex physical poses, and scenarios naturally exhibiting severe mutual occlusion.

\noindent\textbf{Languages and Linguistic Phenomena:} All 2,525 Question-Answer pairs, along with their deterministically generated distractors, are formulated in standard English. The linguistic phenomena evaluated are intentionally constrained to spatial reasoning. The prompts heavily utilize geometric descriptors, spatial prepositions (e.g., ``closest to,'' ``behind,'' ``top-left of''), kinematic pose descriptions, and precise human-object contact relations.

\noindent\textbf{Demographic Representation:} Because SATE-Bench is constructed by repurposing ten established public datasets, the demographic representation inherently mirrors the participant pools of those original source materials. These populations primarily consist of adult athletes, surgical clinicians, and volunteer actors. It is important to note that SATE-Bench is designed exclusively to evaluate 3D geometric reasoning and visual self-awareness; the benchmark does not evaluate, classify, or rely upon demographic traits, facial recognition, or biometric analysis.

\section{AI Usage}
\label{sec:ai_usage}
During the preparation of this manuscript, the authors utilized Large Language Models, such as Gemini and ChatGPT, primarily for draft editing, grammar correction, and prose refinement to enhance the overall readability of the text. Additionally, as detailed in our methodology, MLLMs were employed as an automated fallback mechanism for parsing irregular evaluation outputs and to assist in generating initial distractor options during benchmark curation. All AI-generated text and parsed data are comprehensively reviewed, edited, and validated by the human authors. The core intellectual contributions, experimental designs, human baseline evaluations, and final data annotations are exclusively conceived and executed by the authors.

\section{Answer Extraction and Fallback LLM Usage}
\label{sec:extraction_model_usage}

A rule-based parser extracts the selected option or exploration action from each raw model output. If an output does not follow the single-letter format (e.g., it adds an explanation), an LLM fallback is used to extract the intended decision. 
As shown in Table~\ref{tab:extraction_usage}, all evaluated proprietary MLLMs demonstrated strong instruction adherence capabilities. They achieved a $0.00\%$ fallback usage rate across both the Active multi-turn exploration and the Passive settings, verifying that modern foundation models can strictly abide by formatting constraints even during complex spatial reasoning tasks.

\vspace{-0.5em}
\begin{table}[htbp]
  \centering
  \caption{Usage of the LLM fallback for answer extraction.}
  \label{tab:extraction_usage}
  \resizebox{\linewidth}{!}{
  \begin{tabular}{lrr}
    \toprule
    \textbf{Model} & \textbf{Fallback Used} & \textbf{Usage (\%)} \\
    \midrule
    \multicolumn{3}{l}{\textit{Active}} \\
    Gemini-2.5-flash & 0 & 0.00 \\
    Gemini-2.5-flash-lite & 0 & 0.00 \\
    Gemini-3.1-flash-lite & 0 & 0.00 \\
    Gemini-3.5-flash & 0 & 0.00 \\
    GPT-5-mini-2025-08-07 & 0 & 0.00 \\
    GPT-5-nano-2025-08-07 & 0 & 0.00 \\
    GPT-5.2-2025-12-11 & 0 & 0.00 \\
    \midrule
    \multicolumn{3}{l}{\textit{Passive}} \\
    Gemini-2.5-flash & 0 & 0.00 \\
    Gemini-2.5-flash-lite & 0 & 0.00 \\
    Gemini-3.1-flash-lite & 0 & 0.00 \\
    Gemini-3.5-flash & 0 & 0.00 \\
    GPT-5-mini-2025-08-07 & 0 & 0.00 \\
    GPT-5-nano-2025-08-07 & 0 & 0.00 \\
    GPT-5.2-2025-12-11 & 0 & 0.00 \\
    \bottomrule
  \end{tabular}
  }
\end{table}

\begin{figure*}[t]
\begin{tcolorbox}[
title=Prompts for Active Confidence Score Evaluation,
colback=white,
colframe=black!70,
fonttitle=\scriptsize\bfseries
]
{\footnotesize\textbf{System Prompt.}}\vspace{-0.35em}

\begin{tcolorbox}[
colback=black!2,
colframe=black!10,
boxrule=0.25pt,
arc=0.5mm,
left=0.45mm,
right=0.45mm,
top=0.1mm,
bottom=0.1mm
]
{\scriptsize\ttfamily
\setlength{\baselineskip}{8.2pt}
You are an expert in multi-view humans and objects understanding. 
You are currently viewing 1 out of \{num\_views\} possible camera viewpoints of a single scene. 
These viewpoints are uniformly distributed in a circle around the scene and sorted in clockwise order.\\

Your task is to answer the user's multiple-choice question by carefully analyzing the provided image. 
You must make a strict decision between answering the question or exploring the scene further:\\

1. ANSWER: If the current view provides sufficient information to confidently answer the question, output the correct option.\\
2. EXPLORE: If the current view lacks crucial information, is heavily occluded, or if the question requires full-scene context that this single angle cannot provide, you must request a new viewpoint.\\
\ \ \ - Option 'X': Move to the adjacent left (clockwise) viewpoint.\\
\ \ \ - Option 'Y': Move to the adjacent right (counter-clockwise) viewpoint.\\

CRITICAL CONSTRAINT:\\
For each turn, you must output both your single letter choice (from the set [A, B, C, D, E, X, Y]) and a confidence score between 0.0 and 1.0 (where 1.0 is absolute certainty and 0.0 is complete uncertainty).\\
You must output in the following format:\\
Prediction: <letter>\\
Confidence: <score>\\

Do not include any other explanations, formatting, punctuation, or extra text.
}
\end{tcolorbox}

{\footnotesize\textbf{User Instruction Input Prompt for 5-Option Question.}}\vspace{-0.35em}

\begin{tcolorbox}[
colback=black!2,
colframe=black!10,
boxrule=0.25pt,
arc=0.5mm,
left=0.45mm,
right=0.45mm,
top=0.1mm,
bottom=0.1mm
]
{\scriptsize\ttfamily
\setlength{\baselineskip}{7.5pt}
You are given 1 view or multiple views of the scene.\\

Question: \{question\}\\

Options: A) \{optA\} \; B) \{optB\} \; C) \{optC\} \; D) \{optD\} \; E) \{optE\}\\

Exploration Options: X) \{optX\} \; Y) \{optY\}\\

Please output exactly in this format:\\
Prediction: <letter from [A, B, C, D, E, X, Y]>\\
Confidence: <number from 0 to 1>
}
\end{tcolorbox}

{\footnotesize\textbf{User Instruction Input Prompt for 2-Option Question.}}\vspace{-0.35em}

\begin{tcolorbox}[
colback=black!2,
colframe=black!10,
boxrule=0.25pt,
arc=0.5mm,
left=0.45mm,
right=0.45mm,
top=0.1mm,
bottom=0.1mm
]
{\scriptsize\ttfamily
\setlength{\baselineskip}{7.5pt}
You are given 1 view or multiple views of the scene.\\

Question: \{question\}\\

Options: A) \{optA\} \; B) \{optB\}\\

Exploration Options: X) \{optX\} \; Y) \{optY\}\\

Please output exactly in this format:\\
Prediction: <letter from [A, B, X, Y]>\\
Confidence: <number from 0 to 1>
}
\end{tcolorbox}

\end{tcolorbox}
\caption{Prompts for Active Confidence Score Evaluation}
\label{fig:prompt_active_confidence}
\end{figure*}

\begin{figure*}[t]
\begin{tcolorbox}[
title=Prompt for Active Exhaustive Search Baseline,
colback=white,
colframe=black!70,
fonttitle=\scriptsize\bfseries
]
{\footnotesize\textbf{System Prompt.}}\vspace{-0.35em}

\begin{tcolorbox}[
colback=black!2,
colframe=black!10,
boxrule=0.25pt,
arc=0.5mm,
left=0.45mm,
right=0.45mm,
top=0.1mm,
bottom=0.1mm
]
{\scriptsize\ttfamily
\setlength{\baselineskip}{8.2pt}
You are an expert in multi-view humans and objects understanding. 
You are currently viewing 1 out of \{num\_views\} possible camera viewpoints of a single scene. 
These viewpoints are uniformly distributed in a circle around the scene and sorted in clockwise order.\\

Your task is to answer the user's multiple-choice question. Since this requires an exhaustive understanding of the scene from all angles, you must perform an EXHAUSTIVE SEARCH. 
This means you should explore and view all available viewpoints around the scene before finally choosing to answer the question.\\

1. EXPLORE (highly encouraged until all angles are seen): If you have not yet examined all \{num\_views\} viewpoints of the scene, you must continue to explore:\\
\ \ \ - Option 'X': Move 1 viewpoint to the adjacent left (clockwise).\\
\ \ \ - Option 'Y': Move 1 viewpoint to the adjacent right (counter-clockwise).\\
2. ANSWER: Only when you have completed a thorough, exhaustive search of all angles, or if you are absolutely confident and have reached the maximum exploration limit, output the correct option.\\

CRITICAL CONSTRAINT:\\
Your final output must be exactly ONE single letter from the set [A, B, C, D, E, X, Y]. 
Do not include any explanations, formatting, punctuation, or extra text. Directly output the single letter.
}
\end{tcolorbox}

\end{tcolorbox}
\caption{System Prompt for Active Exhaustive Search Evaluation}
\label{fig:prompt_active_exhaustive}
\end{figure*}